\documentclass[sn-mathphys,Numbered]{sn-jnl}% Math and Physical Sciences Reference Style
\usepackage{adjustbox}

\usepackage{graphicx}%
\usepackage{multirow}%
\usepackage{amsmath,amssymb,amsfonts}%
\usepackage{amsthm}%
\usepackage{mathrsfs}%
\usepackage[title]{appendix}%
\usepackage{xcolor}%
\usepackage{textcomp}%
\usepackage{manyfoot}%
\usepackage{booktabs}%
\usepackage{algorithm}%
\usepackage{algorithmicx}%
\usepackage{algpseudocode}%
\usepackage{listings}%
\usepackage{times}    % Integrate Times for here
\usepackage{xspace}
\usepackage{graphicx}
\usepackage{amsmath}
\usepackage{amssymb}
\usepackage{booktabs}
\usepackage{tikz}
\usetikzlibrary{shapes.geometric, arrows}

\usepackage{array}
\usepackage{rotating}

\usepackage[numbers,sort&compress]{natbib}
\usepackage{silence}  % Suppress unwanted warnings
\usepackage{etoolbox}
\usepackage{float}
\usepackage[format=plain,labelformat=simple,labelsep=period,font=small,compatibility=false]{caption}
\usepackage[font=footnotesize,skip=3pt,subrefformat=parens]{subcaption}
\theoremstyle{plain} % or use `definition` or `remark` based on your needs

\begin{document}

\title[Article Title]{Real-Time Threaded Houbara Detection and Segmentation for Wildlife Conservation using Mobile Platforms}

\author[1]{\fnm{Lyes} \sur{Saad Saoud}}%\email{lyes.saoud@ku.ac.ae}
\author[2]{\fnm{Loïc} \sur{Lesobre}}%\email{loic.lesobre@reneco.org}
\author[2]{\fnm{Enrico} \sur{Sorato}}%\email{enrico.sorato@reneco.org}
% \author[2]{\fnm{Yves} \sur{Hingrat}}%\email{yves.hingrat@reneco.org}
% \author[1]{\fnm{Lakmal} \sur{Seneviratne}} %\email{seneviratne.mudigansalage@ku.ac.ae}
\author[1]{\fnm{Irfan} \sur{Hussain}\thanks{Corresponding author. Email: irfan.hussain@ku.ac.ae}}\email{irfan.hussain@ku.ac.ae}

\affil[1]{\orgdiv{Khalifa University Center for Autonomous Robotic Systems (KUCARS)}, \orgname{Khalifa University}, \orgaddress{\city{Abu Dhabi}, \country{United Arab Emirates}}}
\affil[2]{\orgname{RENECO International Wildlife Consultants LLC},\city{Abu Dhabi},  \country{United Arab Emirates}}

\abstract{Real-time animal detection and segmentation in natural environments play an increasingly important role in wildlife conservation, enabling non-invasive monitoring via remote camera streams. However, these tasks remain challenging due to computational constraints and the cryptic appearance of many animal species. {To address these limitations, we propose a two-stage deep learning framework optimized for mobile platforms, integrating a Threading Detection Model (TDM) to enable parallel processing of YOLOv10-based detection and MobileSAM-based segmentation. Unlike prior YOLO+SAM approaches, our method enhances real-time performance through task parallelization, reducing inference latency.}  Our approach employs YOLOv10 for precise animal detection and MobileSAM for efficient segmentation, leveraging threading techniques to process both tasks concurrently. This optimization significantly improves real-time performance on mobile devices by minimizing delays and optimizing resource utilization.
We evaluate our framework on the Houbara Bustard, a conservation-priority bird species with a cryptic appearance. Our model achieves a mean average precision (mAP50) of 0.9627, mAP75 of 0.7731, and mAP95 of 0.7178 for detection, while MobileSAM attains a mean Intersection over Union (mIoU) of 0.7421 for segmentation. YOLOv10 processes frames in 43.7ms demonstrating its suitability for real-time applications. 
To support our research, we introduce a comprehensive Houbara dataset comprising 40,000 images, addressing a critical gap in conservation AI resources. This dataset covers diverse scenarios, ensuring robust and generalizable model training. Beyond wildlife monitoring, our framework is applicable to real-time object detection and segmentation on resource-constrained devices. By overcoming the challenges associated with cryptic species detection, our work advances computer vision applications in animal behavior analysis and conservation.

}

\keywords{Animal Detection, Mobile Object Segmentation, Wildlife Conservation, Houbara bustard, Computer Vision, Deep Learning, Real-Time Processing, Threading Techniques, Mobile Platforms}

\maketitle
\begin{figure}[t]
    \centering
    \includegraphics[width=1\textwidth]{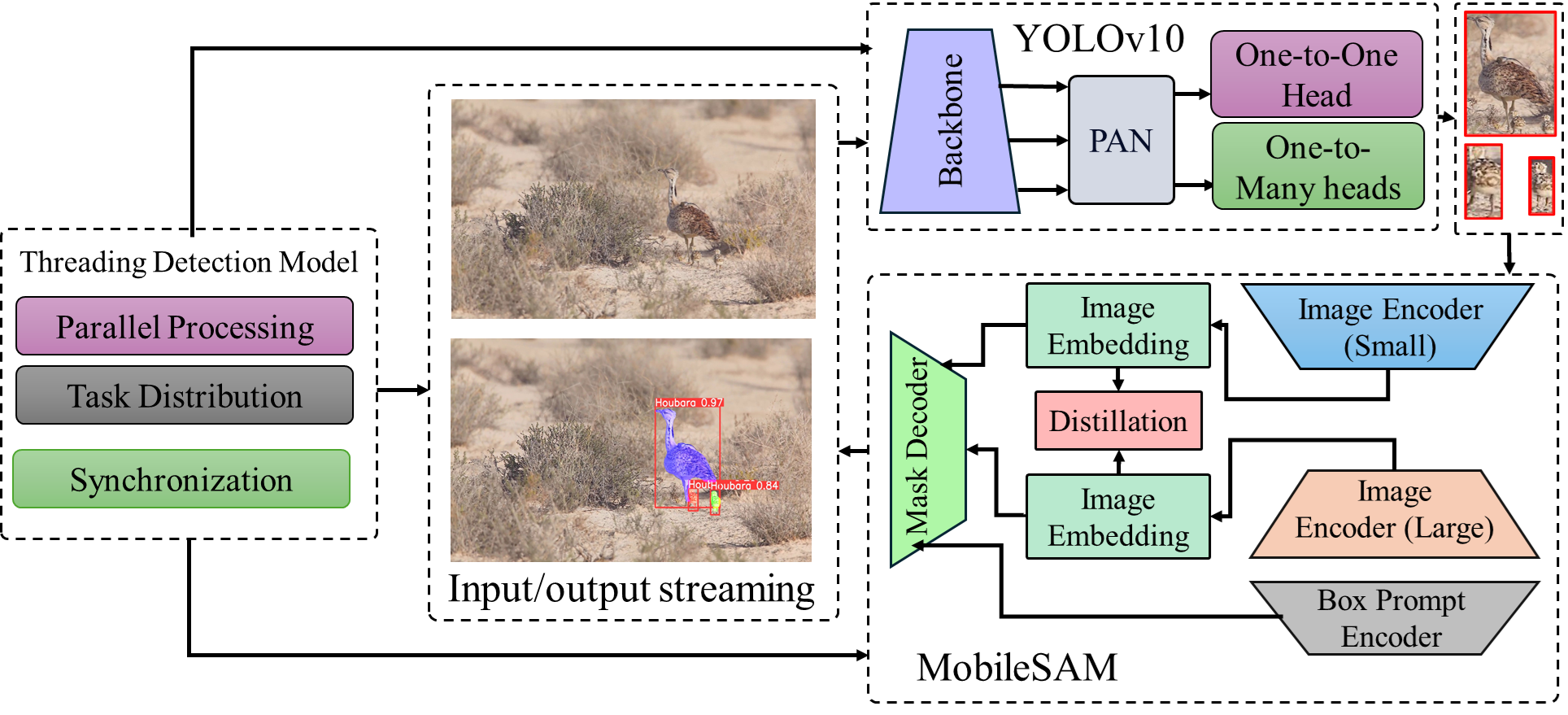}
    \caption{Framework of the Proposed Two-Stage Deep Learning Approach for Houbara Detection and Segmentation, incorporating the Threading Detection Model (TDM). The figure illustrates the end-to-end process starting from input images, through YOLOv10 for improved Houbara detection, followed by MobileSAM for specialized segmentation. The TDM manages parallel processing, task distribution, and synchronization to enhance real-time performance and efficiency throughout the system.}
    \label{fig:main}
\end{figure}

\section{Introduction}\label{sec1}

Advancements in computer vision have significantly improved real-time object detection and segmentation, empowering automated wildlife monitoring and behavioral analysis \cite{Egnor2016, Li2022, Zhang2022, Chen2023}.

% The Houbara Bustard (\textit{Chlamydotis undulata} \& \textit{C. macqueenii}), classified as vulnerable by the International Union for Conservation of Nature (IUCN), faces population declines due to habitat loss and unregulated hunting \cite{birdlife2024asianhoubara, Chopra2024}. Traditional monitoring techniques, such as camera traps and human observation, struggle with inefficiencies due to the species' cryptic nature and vast habitat range \cite{Geary2022, Chilvers2014, Morales2016, Williams2020}.  

To address the challenges of monitoring wild animal populations, real-time detection and segmentation technologies are increasingly integrated into conservation practice \cite{Zhao2024, Sudeepa2024, Wittemyer2019}. Overcoming limitations of classic approaches, real-time monitoring reduces human intervention and facilitates behavioral tracking of species and individuals, aiding conservationists in studying animal behavior, space use and habitat preferences \cite{MacNearney2016}. Real-time monitoring is also increasingly employed for Illegal hunting prevention and habitat disturbance assessment, allowing rapid response to threats faced by animal species of conservation concern \cite{Wittemyer2019}. These technologies are particularly valuable for mobile robotic devices, including aerial drones and mobile terrestrial robots \cite{SAADSAOUD2025102939}, enabling non-invasive, real-time detection and tracking of animals in their natural habitats.

However, despite advances in AI, several challenges persist, arising from Environmental Variability (habitat types, varying light conditions), animal Camouflage, Occlusions of target animals, Computational Constraints under real-time deployment \cite{Sudeepa2024}, and communication Network Limitations in remote conservation areas \cite{Bandaru2024, Zhao2024}.  Deep learning architectures and dataset augmentation strategies for robust detection are needed to overcome some of these challenges.

Here, we propose a two-stage deep learning framework optimized for real-time deployment on mobile platforms to overcome detection challenges and real-time computation constraints. Our approach integrates YOLOv10 for high-accuracy object detection and MobileSAM for instance, segmentation, ensuring both efficiency and accuracy

We developed our real-time tracking system for research applied to monitoring of the Houbara Bustard (Chlamydotis undulata \& C. macqueenii), a bird classified as vulnerable by the International Union for Conservation of Nature (IUCN), facing population declines due to habitat loss and unregulated hunting \cite{birdlife2024asianhoubara, Chopra2024}.
  
{We introduce a 40,000-image dataset specifically curated for Houbara detection and segmentation, filling a critical gap in conserving AI resources. Unlike existing bird detection image data focusing on species classification (e.g.  NABirds), our dataset is tailored for species-specific real-time segmentation and tracking applications. Unlike traditional object detection methods that rely solely on bounding boxes, segmentation provides pixel-level delineation of the bird’s shape, enabling more precise tracking, behavior analysis, and interaction modeling. This fine-grained understanding is particularly crucial in wildlife monitoring, where occlusions, complex backgrounds, and variable postures make simple bounding box detection insufficient. Future work can extend our framework to additional datasets, including other animal species, to further validate its generalization capabilities \cite{Sudeepa2024, Zhao2024}.}

Our contributions advance real-time wildlife monitoring with broader potential applications, including ecological research and autonomous agriculture, particularly in resource-constrained environments where efficient real-time processing is essential.

\section{Motivation}

The demand for real-time wildlife monitoring, especially for cryptic species like the Houbara Bustard, presents challenges for conventional detection methods \cite{Geary2022, Chilvers2014}. Our research optimizes deep learning models for mobile conservation platforms to address these limitations.  

 {\noindent \textbf{Real-Time Detection for Conservation:} 
Traditional methods for monitoring animals in the wild face limitations; for instance, GPS tracking, while allowing continuous monitoring of animal movements over wide areas,  can only indirectly infer the behavior of the animals being tracked; camera traps can capture static images, but are  limited in their capabilities for continuous image processing  \cite{Wittemyer2019, Zhao2024}. Our system can deliver a continuous real-time visual feed with automated detection and segmentation, which can be implemented on mobile robotic systems, offering conservationists a proactive tool for real-time monitoring.
Real-time mobile monitoring can enhance conservation efforts by enabling species tracking in remote locations without onsite human intervention. It can further facilitate behavioral analysis by capturing movement patterns, social interactions, and responses to environmental stimuli. Additionally, it can aid anti-poaching and habitat protection by allowing prompt interventions to threats ensuing from human activities.}  

{\noindent Real-time monitoring enhances conservation efforts by enabling autonomous species tracking in remote locations without human intervention. It facilitates behavioral analysis by capturing movement patterns, social interactions, and responses to environmental stimuli. Additionally, it aids anti-poaching and habitat protection by deploying instant alerts when unauthorized activities are detected, allowing conservation teams to respond swiftly.}  

\textbf{Computational Constraints and Network Latency:} Mobile platforms have limited processing power, restricting real-time performance.  
{\noindent In remote environments with limited power and connectivity, cloud-based solutions are impractical, requiring on-device inference. Our approach optimizes deep learning models for deployment on platforms like NVIDIA Jetson Xavier, ensuring real-time processing without constant internet access.}  

\textbf{Challenges in Detecting Cryptic Species:} The Houbara Bustard’s natural camouflage and stillness make detection difficult \cite{Hending2024, Humphrey2017}.  
{\noindent Unlike general object detection, cryptic species require models that capture subtle textures and contextual cues. Shadows, rocks, and vegetation introduce background noise, leading to misclassification \cite{Karp2020, Gonzalez2009}. Our approach integrates adaptive augmentations, exposing models to varying lighting, occlusions, and distractors for improved robustness.}  
{\noindent We further enhance detection with context-aware segmentation, leveraging spatial and behavioral cues beyond visual contrast. The dataset includes challenging samples with occlusions and lighting variations, ensuring real-world robustness.}  

\textbf{Lack of Suitable Datasets:} High-quality datasets for small-to-medium-sized species remain limited, affecting model generalization.  
{\noindent While datasets like CUB-200 and NABirds focus on image-level classification, they lack fine-grained instance segmentation and real-time tracking crucial for conservation applications. Our dataset fills this gap by emphasizing instance segmentation, real-time tracking, and behavioral analysis. Future work will explore integrating and benchmarking against datasets like CUB-200.}  

\section{Contributions}

This paper presents key contributions to real-time wildlife monitoring using computer vision, specifically for the detection and segmentation of the Houbara Bustard:

\begin{itemize}
   \item {\textbf{Threaded Real-Time Deep Learning Framework:} We introduce a two-stage deep learning framework optimized for mobile platforms, integrating the Threading Detection Model (TDM) to enable parallel execution of YOLOv10 for detection and MobileSAM for segmentation. Unlike prior YOLO-SAM approaches, our framework improves inference efficiency by reducing latency and optimizing resource utilization. TDM is model agnostic, allowing seamless integration with alternative object detectors (e.g., YOLOv8, EfficientDet) and segmentation models (e.g., HQ-SAM, DeepLabV3).}  

   \item \textbf{Efficient Real-Time Processing with TDM:} The Threading Detection Model (TDM) parallelizes detection and segmentation while optimizing network transmission, reducing delays in remote monitoring. {By integrating low-latency threading techniques and customizable multi-model pipelines, our system supports applications such as automated population monitoring, anti-poaching alerts, and real-time behavior tracking.}

   \item  \textbf{Comprehensive Houbara Bustard Dataset:} We introduce a high-quality dataset of 40,000 annotated images for houbara detection and segmentation. This dataset addresses a critical gap in wildlife conservation research by providing diverse and well-annotated images, facilitating the development of robust deep-learning models for ecological studies.  

   \item {\textbf{Generalization and Scalability Beyond Houbara Bustards:} Unlike existing datasets that focus on image classification (e.g., NABirds), our dataset is tailored for real-time instance segmentation and behavior tracking. While curated for Houbara research and conservation, our framework is adaptable to other species and supports adaptive segmentation strategies to handle occlusion, environmental variability, and motion blur. Future work will explore its application to additional avian datasets and large-scale wildlife monitoring systems.}  
\end{itemize}

\section{Literature review}

\subsection{Wildlife Detection and Camouflage Challenges}

Detecting wildlife in natural environments is challenging, especially for cryptic species like the Houbara Bustard. Detection models must address occlusions, background complexity, and environmental variability. Traditional methods, such as thresholding and color-based segmentation \cite{Xiao2016}, struggle in dynamic conditions where lighting changes and textures obscure object boundaries.

{
Recent deep-learning advancements have improved camouflage detection through spectral filtering, motion-based tracking, and attention-driven models \cite{Mpouziotas2024, Ma2024}. Techniques such as multispectral imaging and CBAM-augmented YOLO networks enhance object-background separation, while optical flow analysis captures subtle motion cues for better tracking \cite{Rahman2021, Blanchette2017}.}

Modern deep learning models, including Faster R-CNN, MobileNet-v3, RetinaNet, and YOLO variants, achieve high accuracy in wildlife detection but still struggle with small object recognition, dataset imbalance, and cryptic species differentiation \cite{song2024, xu2024}. Hybrid methods, like glance-and-stare frameworks, integrate deep learning with classical image processing, e.g. to improve localization in aerial monitoring \cite{tian2019}. UAVs equipped with thermal imaging enhance nocturnal and long-range detection, while AI-powered filtering improves camera trap data by reducing false positives \cite{Rahman2021, Burton2015}.

Despite advancements, detecting camouflaged animals remains difficult due to background complexity, textural similarities, and behavioral adaptations that reduce visibility \cite{Dyer2024}. Real-time deployment on embedded AI devices further necessitates model optimization for computational efficiency in remote conservation applications \cite{Mpouziotas2024}.

{Future research should integrate multimodal approaches, such as spectral imaging, motion tracking, and real-time segmentation, to improve wildlife monitoring. Advances in deep learning and edge AI will further enhance camouflage detection, supporting wildlife research and conservation efforts for species like the Houbara Bustard.}

\subsection{Comparison/overview of  animal Detection Methods in natural environments for Houbara Bustard}

In wildlife studies employing image processing techniques, selecting the most effective algorithms is crucial for achieving accurate and reliable results. Traditional methods for image analysis and deep learning approaches offer distinct advantages and face unique challenges. Classic machine learning-based methods often rely on processes such as binarization and denoising  \cite{Simon2020}. However, these techniques face significant obstacles due to the similarities in color and texture between the studied animals and their surroundings, which can adversely impact detection accuracy \cite{Wang2023}. Moreover, traditional machine learning-based methods generally have limited feature capture capabilities, leading to reduced detection accuracy and robustness, and often consume substantial resources  \cite{Wu2022}.

In contrast, deep learning techniques have emerged as powerful alternatives, offering significant improvements in detection performance and enabling on-the-fly animal segmentation and identification. DL methods can identify various animal species by leveraging robust feature extraction capabilities, enhancing accuracy and reliability \cite{Mu2023}. Convolutional neural networks (CNNs), in particular, provide strong support for detection tasks, demonstrating higher performance than traditional methods \cite{Yu2021}. Nonetheless, deep learning approaches also face challenges, such as detecting small-size subjects and handling cryptic protective coloration. In addition, these methods require time-consuming expert annotation for training and must be optimized to run efficiently on lightweight devices \cite{Yuan2022}.

Overall, deep learning methods generally outperform traditional approaches in terms of accuracy and reliability. While classic ML methods require additional preprocessing and often struggle with environmental factors affecting accuracy \cite{Simon2020}, deep learning techniques offer enhanced feature extraction capabilities, resulting in superior performance \cite{Wang2023} \cite{Yuan2022}. A hybrid approach, integrating both traditional and deep learning techniques, may offer a balanced solution by leveraging the strengths of both methodologies while addressing their respective limitations \cite{Wang2023} \cite{Mu2023} \cite{Yuan2022}.

\subsection{Real-Time Object Detection and Segmentation on Mobile Devices}

Developing real-time object detection and segmentation algorithms for mobile devices presents challenges due to computational, power, and performance constraints \cite{Lee2023, Jose2019}.  {Recent advancements in conservation AI have explored real-time tracking applications, including UAV-assisted wildlife monitoring, thermal-imaging-based animal detection, and bioacoustic signal integration with real-time image recognition \cite{Zhao2024, Scholz2016}. Studies indicate that achieving at least 15 FPS on embedded devices is crucial for seamless real-time processing in conservation tools  \cite{He2024}.} To address these constraints, lightweight deep learning models such as Binarized Neural Networks (BNNs) and MobileDenseNet have been optimized for embedded systems, utilizing model compression and quantization to enhance efficiency without sacrificing accuracy \cite{Hajizadeh2023, Cai2021}. Hardware-accelerated approaches, including FPGA implementations and GPU-CPU collaborative schemes, further improve real-time performance on resource-limited devices \cite{Yu2018, Cai2021}.

Deploying object detection and segmentation models on mobile devices further requires strategies to overcome limited computational resources, memory constraints, and energy consumption. Techniques such as model pruning, quantization, and specialized hardware accelerators have been employed to optimize inference efficiency \cite{praneeth2023scaling, chebotareva2024comparative}. For example, lightweight feature extractors like ResNet18 have demonstrated high inference efficiency while maintaining competitive accuracy, even on single-threaded CPU execution \cite{bai2019efficient, li2021depthwise, li2020mspnet}. Additionally, algorithm-hardware co-optimization, such as integrating binarized neural networks with FPGA-based detection processors, leverages parallel computing for higher throughput on mobile devices \cite{lee2023real}.

 Implementations on platforms like Raspberry Pi and NVIDIA Jetson Nano provide insights into trade-offs between accuracy, speed, and computational efficiency, emphasizing the need to balance resource constraints with real-time performance  \cite{chebotareva2024comparative}.

{
Prior research has explored YOLO + SAM integrations for various applications. Zhang et al. \cite{zhang2023fastersam} optimized SAM for mobile use but did not integrate it with an object detection model. Pandey et al. \cite{pandey2023multimodal}  combined YOLOv8 with SAM for multimodal medical segmentation but did not address real-time inference for mobile and edge devices. Our approach introduces the Threading Detection Model (TDM), a novel framework enabling parallel execution of YOLOv10-based detection and MobileSAM-based segmentation. By leveraging multi-threaded processing, TDM reduces inference latency and optimizes computational resources, making real-time segmentation feasible for conservation applications. This improvement is essential for tracking cryptic species in dynamic environments, where low-latency detection and segmentation are critical yet constrained by hardware limitations.
}

\subsection{Wildlife Monitoring Applications Using Computer Vision}

Computer vision plays an increasingly crucial role in wildlife research and monitoring, advancing species identification, tracking, and behavioral analysis. These techniques enhance ecological research and conservation efforts by enabling automated detection and tracking of animal species in diverse habitats. However, challenges such as dataset limitations, environmental variability, and real-time processing constraints must be addressed to ensure the effectiveness of these tools.

A key limitation in wildlife monitoring is the lack of robust datasets tailored for tracking wild animals. Existing datasets primarily focus on controlled environments, limiting their applicability in the wild. To address this, the Wild Animal Tracking Benchmark (WATB) was introduced to foster research on visual object tracking across diverse species \cite{wang2023watb}. This benchmark promotes the development of models capable of adapting to the variability of wildlife settings.

Another challenge arises in drone-based bird detection, where dynamic flight movements and long-range imaging introduce complexities in capturing clear and consistent data. Advanced models like ORACLE are designed to handle such constraints, improving detection and tracking accuracy in aerial footage \cite{mpouziotas2024advanced}. Similarly, monitoring small to medium-sized animals is hindered by a lack of annotated datasets. The Small-sized Animal Wild Image Dataset (SAWIT) addresses this gap by providing essential resources for ecological studies \cite{nguyen2024sawit}.

{\noindent \textbf{Real-Time Tracking and Multi-Modal Approaches:} 
Beyond static object detection, wildlife monitoring increasingly integrates real-time tracking with multi-sensor approaches, including RGB, UVand infrared imaging. For instance, Camera trap-based surveillance uses motion sensors and night-vision capabilities for passive monitoring in low-light conditions \cite{Sudeepa2024}. UAV-based tracking leverages deep learning models for species identification and movement tracking from aerial imagery, including RGB and IR channels \cite{Zhao2024}.}

{\noindent \textbf{2D and 3D Animal Tracking Frameworks:} 
Traditional 2D object detection techniques analyze individual frames, providing position information within a fixed image plane. However, these methods lack depth estimation and are limited in tracking animals that move beyond a single plane (e.g., birds in flight or swimming fish). In contrast, 3D tracking incorporates depth estimation and motion trajectories, allowing for accurate localization in three-dimensional space, which is crucial for animals exhibiting significant vertical movement. Furthermore, thermal and infrared tracking models improve visibility under challenging conditions, such as nocturnal activity or occlusions, mitigating the limitations of conventional RGB-based detection \cite{He2024}.}

% While 2D object detection techniques analyze single frames, 3D tracking methods improve spatial awareness by estimating depth and movement trajectories, enhancing localization in cluttered environments \cite{He2024}. Additionally, thermal and infrared tracking models provide better visibility in nocturnal and occluded conditions, addressing limitations of conventional RGB-based detection.}

Despite these advancements, challenges such as background clutter, varying lighting, and multi-species co-occurrence complicate animal recognition and tracking \cite{islam2020herpetofauna}. Ongoing research must refine computer vision models to improve species differentiation and behavioral analysis in complex environments.
Overall, while computer vision has significantly enhanced wildlife monitoring, continued improvements in dataset diversity, real-time tracking, and multi-modal integration is needed for expanding the scope and effectiveness of wildlife research and conservation.

\section{Houbara Image Dataset}

\noindent \textbf{Objective:} 
The Houbara Dataset is designed to advance deep learning models for detecting and segmenting the Houbara bustard, a cryptic species that poses significant challenges for automated tracking. It provides a diverse, high-quality resource for training and evaluating models in real-world conservation applications.

\noindent \textbf{Dataset Construction:} The dataset integrates in-house recordings, and publicly available sources to ensure a broad representation of poses, lighting conditions, and behaviors. {Camera traps, deployed in varied habitats (deserts, shrublands, steppes), further captured species activity across different ecological settings and time periods (day, night, dawn, dusk).}

\noindent \textbf{Dataset Composition:} 
The dataset consists of 40,000 images, split into 32,000 for training (80\%), 4,000 for validation (10\%), and 4,000 for testing (10\%).   {This distribution ensures balanced and effective model training, validation, and evaluation. The dataset comprises images collected from various sources, including The dataset comprises images collected from various sources, including in-house recording (55\%), camera traps in the wild (18\%), and online sources (27\%)}.

{To ensure a fair evaluation process, the test set mirrors the proportional distribution of images across different sources, mitigating the risk of domain bias and improving model robustness. Additionally, we analysed key environmental challenges within the dataset, such as lighting variations, motion blur, and object occlusions. The categorization of images into these conditions was conducted using a set of predefined thresholds on key image properties, including brightness levels, blur sharpness, object size, and occlusions. Specifically, the following criteria were used:}
\begin{itemize}
    \item {\textbf{Blurred Images:} Classified based on a Laplacian variance threshold (BLUR\_THRESHOLD = 50), where images below this value were considered blurred.}
    \item {\textbf{Lighting Conditions:} Images were categorized into underexposed, overexposed, or normal based on mean pixel intensity (DARK\_THRESHOLD = 40, BRIGHT\_THRESHOLD = 200.)}
    \item {\textbf{Small Object Detection:} Objects covering less than 2\% of the image area were categorized under small object cases (SMALL\_OBJECT\_THRESHOLD = 0.02).}
    \item {\textbf{Occluded Objects:} High intersection-over-union (IoU \> 0.5) between ground-truth bounding boxes was used to identify occlusions.}
\end{itemize}

{Furthermore, the dataset prioritizes low-light conditions, with 92.57\% of images captured during the day, 6.48\% at dusk/dawn, and 0.95\% at nighttime. Additionally, 12.71\% of images exhibit motion blur, primarily occurring in dusk/dawn scenarios. These characteristics closely replicate real-world conservation challenges, enhancing the model’s adaptability to varying environmental conditions. By implementing systematic classification criteria, our methodology ensures a robust evaluation of the model’s performance under diverse challenges.}

\noindent \textbf{Dataset Splitting Strategy \& Avoiding Data Leakage:} {To prevent data leakage, dataset splits were assigned at the video level, ensuring that frames from the same sequence do not appear in multiple subsets. Camera trap data was separated based on non-overlapping time intervals to ensure that consecutive frames of the same scene do not appear across training, validation, and testing sets. Similarly, in-house recordings were cross-referenced using unique bird identities, ensuring that the same individual bird did not appear in different dataset splits. This approach prevents duplicate appearances, mitigates potential bias, and ensures an unbiased model evaluation without artificially inflating accuracy.}

\noindent \textbf{Image Characteristics:} {The dataset includes images captured using RGB and infrared, from camera traps, for low-light tracking.}

\begin{figure*}[t] 
\centering 
\includegraphics[width=1\textwidth]{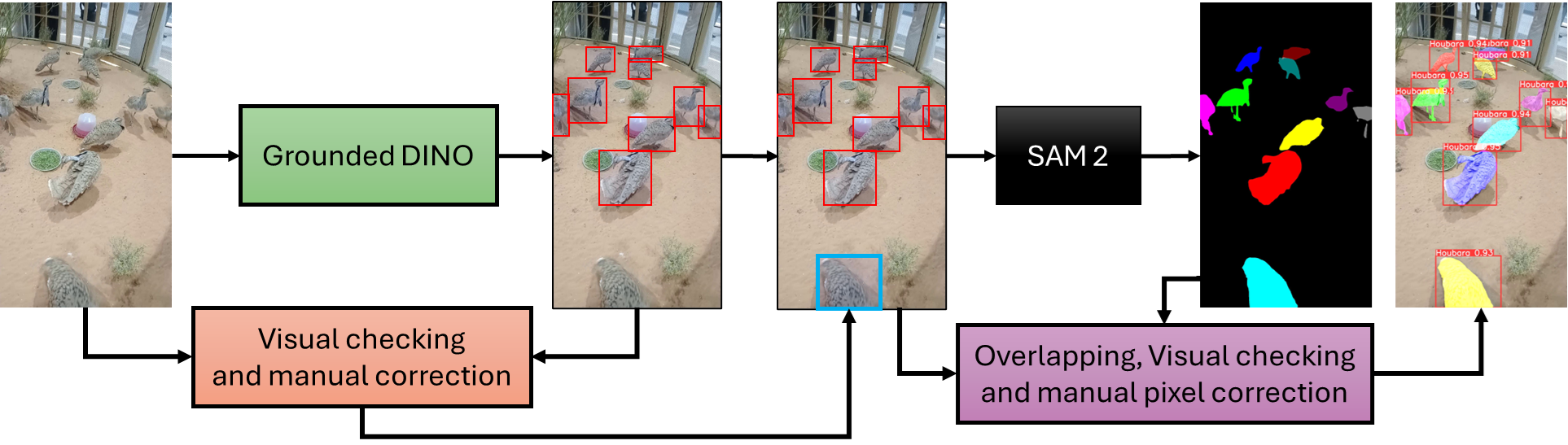}
\caption{Annotation Workflow: Bounding boxes generated with GroundingDINO (Stage 1) followed by detailed segmentation using SAM2 (Stage 2). Manual validation ensures annotation accuracy.} 
\label{fig:ann_process}
\end{figure*}

\noindent \textbf{Annotation Process:} {
Bounding boxes were generated using GroundingDINO, followed by fine-grained segmentation using Segment Anything Model v2 (SAM2). This semi-automated workflow significantly reduced annotation time while maintaining accuracy, with manual refinement applied to 12\% of cases, particularly for images with occlusions or challenging lighting conditions. 
The average inference time per image was {0.1241 seconds} for GroundingDINO and 0.3297 seconds for SAM2, ensuring efficient processing while achieving high segmentation precision. The annotation process was executed on a workstation equipped with an {Intel Core i9-13900K} (12 physical, 20 logical cores) and an NVIDIA RTX 4090 GPU with {24GB VRAM}, leveraging CUDA acceleration to optimize performance. Figure ~\ref{fig:ann_process} illustrates the annotation pipeline.}

\begin{figure*}[t]
    \centering
    \begin{subfigure}[b]{0.3\textwidth}
        \centering
        \includegraphics[width=\textwidth]{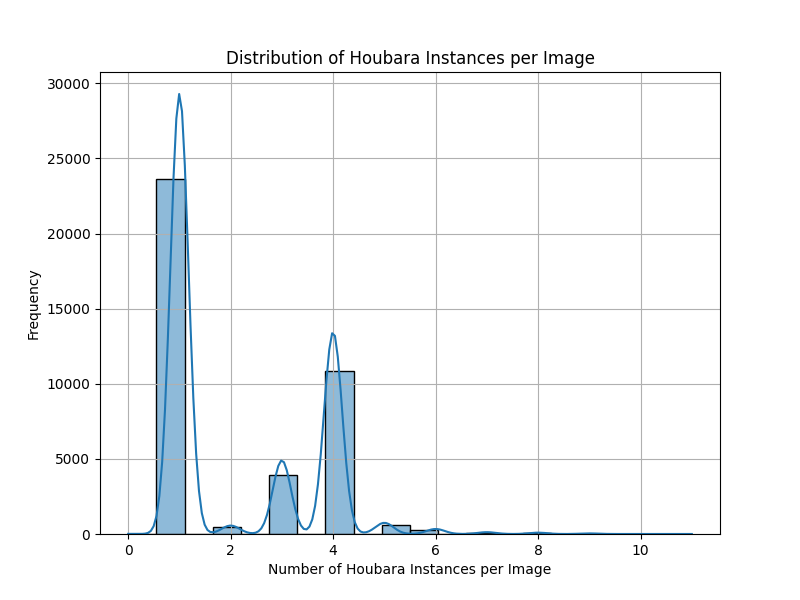}
        \caption{Instance Distribution}
    \end{subfigure}
    \hfill
    \begin{subfigure}[b]{0.3\textwidth}
        \centering
        \includegraphics[width=\textwidth]{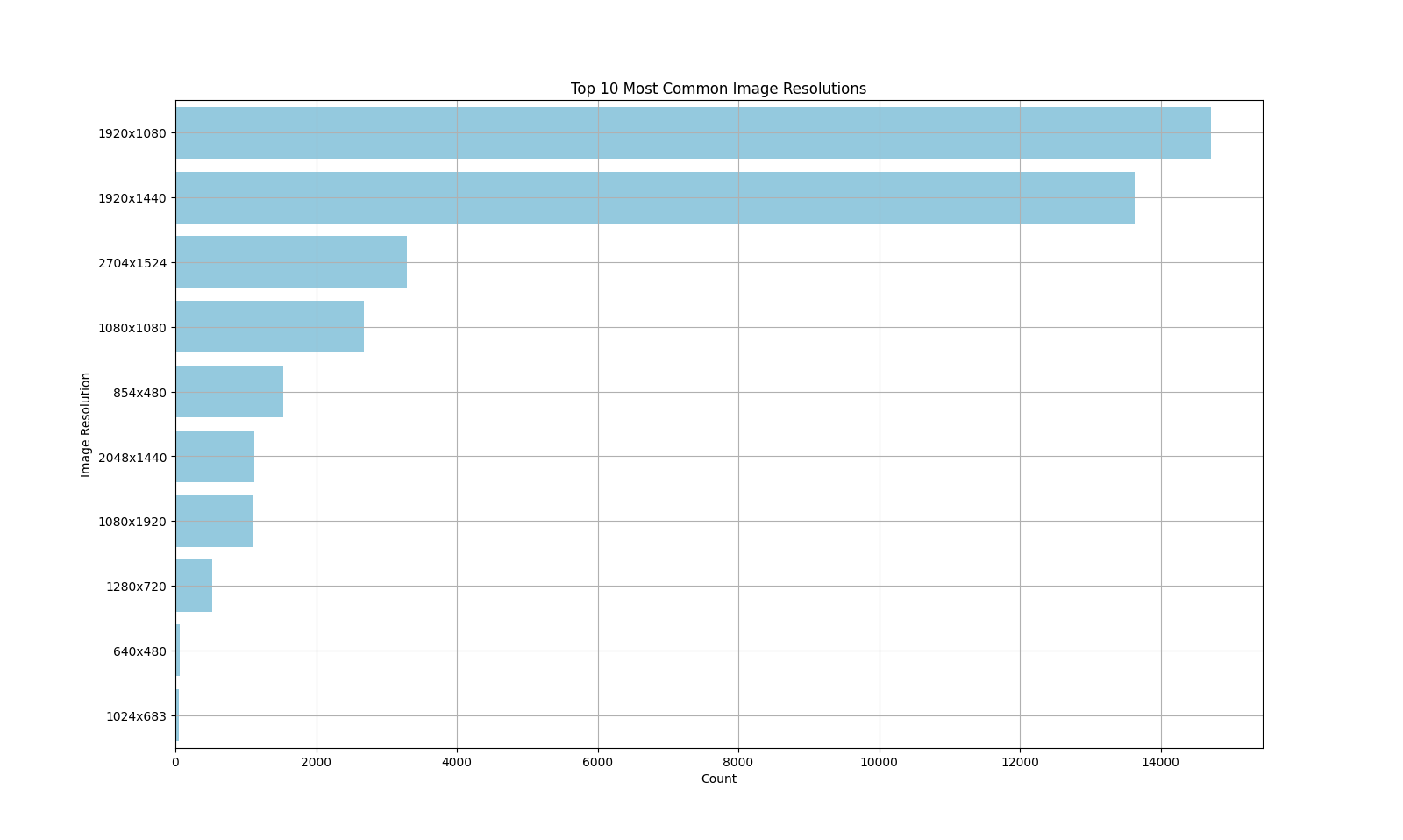}
        \caption{Resolution Distribution}
    \end{subfigure}
    \hfill
    \begin{subfigure}[b]{0.35\textwidth}
        \centering
        \includegraphics[width=\textwidth]{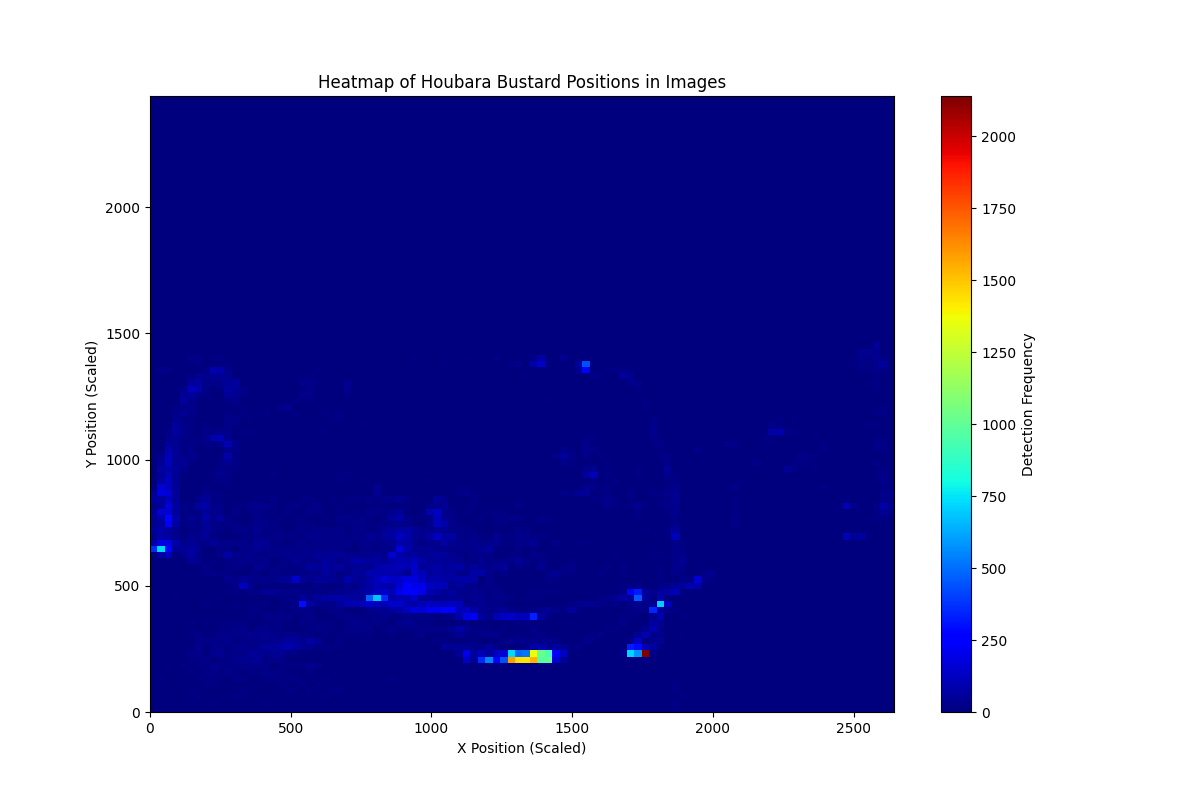}
        \caption{Spatial Heatmap}
    \end{subfigure}
    \caption{Dataset analysis: (a) Number of instances per image, (b) Image resolution distribution, and (c) Spatial heatmap of Houbara positions. These insights ensure dataset balance and diversity.}
    \label{fig:dataset_analysis}
\end{figure*}

\noindent \textbf{Dataset Analysis:} {Figure~\ref{fig:dataset_analysis} provides key insights, including (i) Instance Distribution: A mix of single-bird and multi-bird images ensures object density variation for robust model training. (ii) Image Resolution Distribution: The dataset contains a variety of resolutions, with the most common being 1920×1080 and 1920×1440, ensuring robustness against image quality degradation. (iii) Spatial Distribution Heatmap: Highlights the frequency of Houbara appearances in different image regions, preventing positional biases.}

\noindent \textbf{Comparison with Existing Datasets:} {Unlike bird detection datasets such as CUB-200 and NABirds, which focus on species classification, the Houbara Dataset is designed for real-time single-species segmentation and tracking filling a gap in species-specific wildlife monitoring}

\noindent \textbf{Future Dataset Expansion:} {The dataset will be expanded to 50,000 images, incorporating new geographic locations and seasonal variation. Long-term tracking with camera traps will further refine model adaptation to changing environmental conditions.}

\noindent  {Overall, the Houbara Dataset provides a high-quality, structured resource for advancing real-time wildlife detection and segmentation. Its diversity, robust annotation process, and real-world challenges make it well-suited for conservation applications and future AI research in ecological monitoring.}

\section{Methodology}
\label{sec:methodology}

This section presents our two-stage deep learning approach for efficient wildlife detection and segmentation on mobile platforms (Figure~\ref{fig:main}).

\subsection{Stage 1: Efficient Detection with YOLOv10}

The YOLOv10 architecture  \cite{wang2024yolov10realtimeendtoendobject}  offers significant improvements in computational efficiency and detection accuracy, making it well-suited for real-time wildlife monitoring. {Compared to prior YOLO versions, YOLOv10 enhances feature extraction, refines its detection head, and optimizes processing speed, enabling real-time inference on edge devices like the NVIDIA Jetson AGX Xavier \cite{Scholz2016}. These advancements improve detection precision while reducing computational overhead, ensuring suitability for mobile platforms and conservation applications  \cite{Zhao2024, Sudeepa2024}.}

{Unlike previous approaches that integrate YOLO with SAM for sequential object detection and segmentation \cite{zhang2023fastersam, pandey2023multimodal}, our method introduces the {Threading Detection Model (TDM)}, enabling parallel execution of detection and segmentation. This concurrent processing reduces inference latency, optimizing performance for mobile and resource-constrained environments \cite{He2024}. The ability to track elusive species in dynamic settings with minimal delay is critical for real-time conservation monitoring.} 

Performance benchmarks show that YOLOv10 achieves higher mean average precision (mAP) with lower latency compared to prior YOLO models, making it effective in detecting and tracking fast-moving objects such as the Houbara Bustard \cite{Rahman2021}. {These enhancements ensure real-time wildlife tracking without sacrificing accuracy, making YOLOv10 a preferred choice for conservation applications that demand high-speed processing \cite{MacNearney2016, Valderrama2024}.} 

\subsection{Stage 2: Segmentation with MobileSAM}

For instance segmentation, we utilize MobileSAM, a lightweight segmentation model designed for mobile platforms. It operates within the regions identified by YOLOv10 detections.

\noindent \textbf{MobileSAM Architecture:}  
MobileSAM employs a compact Vision Transformer (ViT) as its image encoder, optimized for efficiency on edge devices. The model generates high-dimensional embeddings, capturing essential image features while maintaining low computational overhead. A key component is its prompt-guided mask decoder, which refines segmentation based on user-defined inputs such as bounding boxes. {This targeted segmentation approach ensures precise mask generation while reducing unnecessary computations, making it highly efficient for mobile deployment.}

\begin{figure}[t]
    \centering
    \includegraphics[width=1\textwidth]{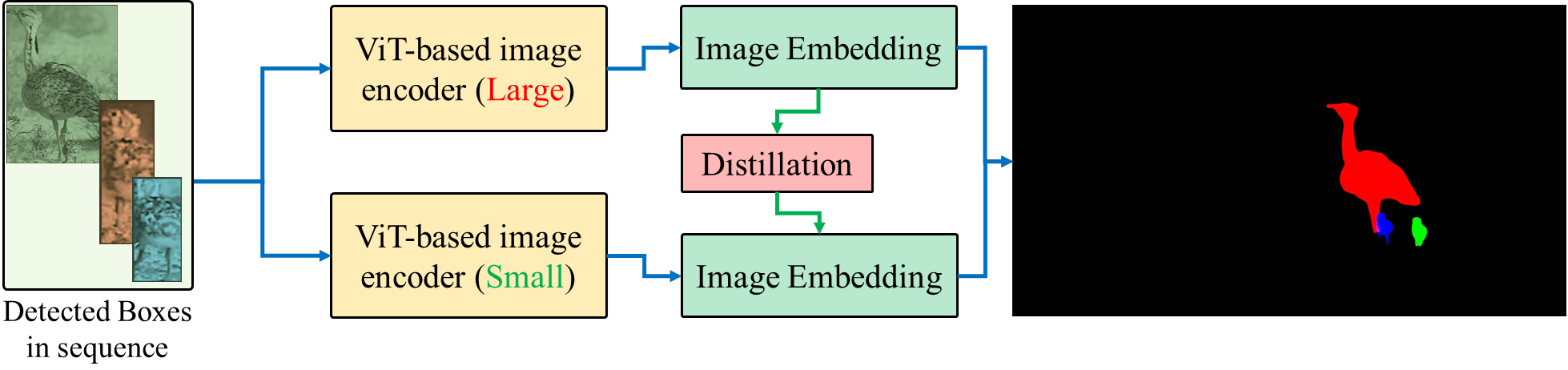}
    \caption{Proposed MobileSAM for Houbara Bustard segmentation. YOLOv10 detects objects and generates bounding boxes, which are then passed to MobileSAM for precise segmentation.}
    \label{fig:HoubaraMobieSAM}
\end{figure}

\noindent \textbf{Advancements Beyond Standard ROI-Based Segmentation:}  
{Traditional top-down segmentation approaches operate sequentially, leading to computational inefficiencies. Our method introduces {TDM}, which allows parallel execution of YOLOv10 detection and MobileSAM segmentation, reducing inference latency and improving resource utilization.}  

{Unlike baseline MobileSAM, which processes entire images, our approach first detects objects using YOLOv10 and applies segmentation only within predefined Regions of Interest (RoIs). This significantly reduces computational load, improves inference speed, and minimizes irrelevant background information.}  

{By leveraging multi-threaded execution, TDM further optimizes hardware utilization, making our framework highly efficient for real-time conservation monitoring. While YOLOv10 provides rapid localization, MobileSAM ensures fine-grained segmentation, improving overall accuracy. This structured approach aligns with existing detection-assisted segmentation frameworks and is validated in Figure~\ref{fig:HoubaraMobieSAM}.}

\subsection{Challenges and Optimization for Mobile Platforms}

Deploying YOLOv10 and MobileSAM on mobile platforms presents multiple challenges that must be addressed for efficient real-time video analysis.

\noindent \textbf{Challenges:}  
- Computational constraints: YOLOv10 demands significant processing power, which can strain mobile hardware.  
- Model accuracy: Occlusions, scale variations, and low-light conditions can degrade detection performance.  
- Network latency: Transmission delays over Wi-Fi can hinder real-time responses, affecting conservation monitoring.  

\noindent \textbf{Optimization via Threaded Parallel Processing:}  
To overcome these challenges, we implemented an optimization strategy that combines fine-tuning with advanced threading techniques. YOLOv10 was fine-tuned for Houbara Bustard detection using species-specific augmentations.

{\noindent The \textbf{TDM architecture} consists of four threads: a video processing thread, a YOLOv10 detection thread, a MobileSAM segmentation thread, and a post-processing thread. Synchronization mechanisms, such as locks and semaphores, ensure seamless frame processing.}  

{By executing detection and segmentation concurrently, TDM minimizes processing delays, ensuring real-time wildlife tracking. Additionally, it optimizes network transmission, mitigating latency issues associated with remote streaming from mobile platforms.}  

\begin{figure*}[t]
  \centering
  \includegraphics[width=\linewidth]{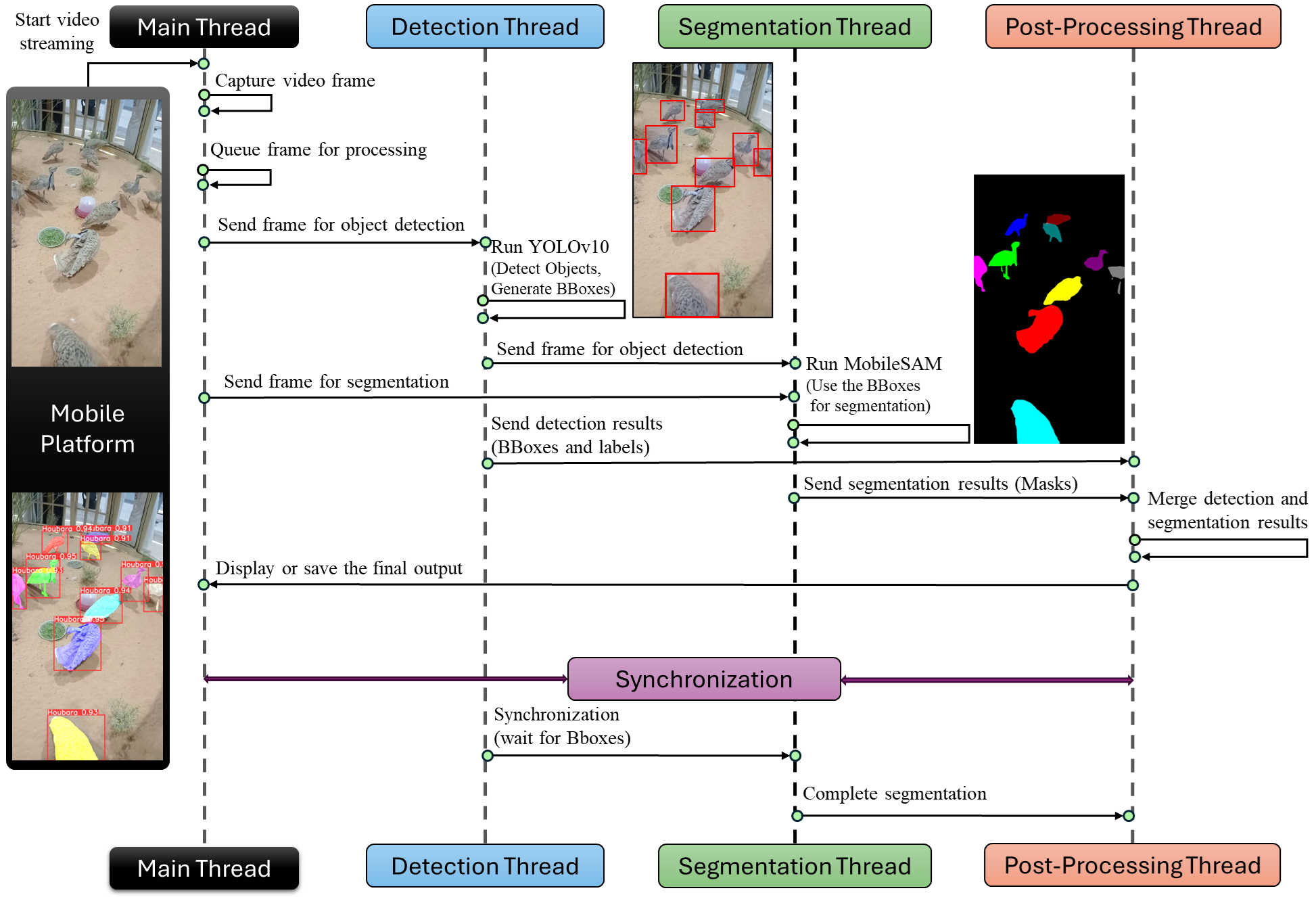}
  \caption{Threaded architecture for real-time object detection and segmentation. The video feed is split into two threads: one for YOLOv10 object detection and the other for MobileSAM segmentation. Results are merged in post-processing for display or storage.}
  \label{fig:threading_architecture}
\end{figure*}

Figure~\ref{fig:threading_architecture} illustrates the TDM framework, demonstrating how concurrent execution significantly improves performance over sequential processing.

\subsection{Training and Deployment Hardware}

{\noindent \textbf{Training Setup:}  
YOLOv10 was trained on an {NVIDIA A100 GPU (40GB VRAM)} using {CUDA 11.8} and {PyTorch 2.2.1}. Training on the 40,000-image dataset took {{14} hours} for convergence over {200 epochs}, with an initial learning rate of {0.01}, momentum of {0.9}, and weight decay of {0.0005}. A {cosine annealing learning rate scheduler} dynamically adjusted learning rates. Augmentation techniques included Mosaic Augmentation, MixUp Augmentation, CLAHE, Random Flipping, and Color Jittering, enhancing model robustness across diverse conditions.}

{\noindent \textbf{Deployment Setup for Real-Time Inference:}  
For real-time inference, the model was deployed on an {NVIDIA Jetson AGX Xavier}, an embedded AI platform optimized for edge computing. Model quantization, TensorRT acceleration, and FP16 precision inference enabled {real-time processing at {12} FPS}. The integration of TDM further reduced inference latency, ensuring seamless operation for conservation monitoring.}

{\noindent \textbf{Computational Cost Considerations:}  
While training requires high-performance GPUs, the inference pipeline is optimized for mobile deployment. The system effectively balances accuracy and efficiency, making it deployable in real-world conservation scenarios. Trade-offs between computational cost and real-time performance were carefully managed to enable effective deployment on resource-constrained devices.}

\section{Testing Setup}
\label{sec:exp_setup}

Experiments were conducted on an NVIDIA Jetson AGX Xavier platform~\cite{jetson_agx_xavier}, selected for its high-performance GPU and memory capabilities, making it suitable for real-time object detection and segmentation in resource-constrained environments. The testing setup ensured rigorous evaluation of accuracy, inference speed, and computational efficiency for embedded AI applications.

\subsection{Object Detection Experiments}
For object detection, we fine-tuned several pre-trained models, including Faster R-CNN~\cite{Girshick_2015_ICCV}, MobileNet-v3~\cite{Howard_2019_ICCV}, RetinaNet~\cite{Lin_2017_ICCV}, FCOS~\cite{9229517}, EfficientDet~\cite{tan2020EffDet}, RTDETR~\cite{lv2023detrs, lv2024rtdetrv2improvedbaselinebagoffreebies}, YOLOv5~\cite{yolov5}, YOLOv8~\cite{reis2023realtime}, and YOLOv10-Medium~\cite{wang2024yolov10realtimeendtoendobject}. Training was performed using PyTorch 2.2.1 with CUDA 11.8 for 100 epochs, with a batch size of 16 and an image resolution of 640 × 640 pixels. 

{\noindent \textbf{YOLOv10 Training Setup:}  
YOLOv10-Medium was selected for its balance between computational efficiency and detection accuracy, making it ideal for real-time applications. Training utilized the SGD optimizer with an initial learning rate of 0.01, momentum of 0.9, and a cosine annealing scheduler for dynamic rate adjustment. Data augmentation techniques applied using Albumentations, a widely used image augmentation library to improve model robustness, included {Blur} (p=0.01), {MedianBlur} (p=0.01), {ToGray} (p=0.01), and {Contrast Limited Adaptive Histogram Equalization (CLAHE)} (p=0.01). Additional augmentations such as {Mosaic Augmentation}, {MixUp Augmentation}, {Random Flipping}, {Color Jittering}, and {Scaling \& Cropping} further improved robustness. The dataset validation process confirmed {4,000 images}, ensuring integrity by removing background-only samples or corrupt files. TensorBoard was enabled for real-time monitoring of training progress.}

\subsection{Segmentation Experiments}
For segmentation, we evaluated DeepLabV3~\cite{chen2017rethinking}, Lite R-ASPP (LRASPP)~\cite{Howard_2019_ICCV}, FCN-ResNet-50~\cite{long2015fully}, and MobileSAM-based approaches. Models were trained for 100 epochs with a batch size of 8 using the Adam optimizer (learning rate 0.0005). 

{\noindent \textbf{Comparison Between Prompt-Based and Non-Prompt-Based Models:}  
While MobileSAM and SAM rely on prompt-based segmentation, conventional models (DeepLabV3, LRASPP, FCN) employ direct segmentation. Evaluating both paradigms highlights the trade-offs in computational efficiency, segmentation quality, and real-time applicability for conservation monitoring. A standardized dataset split ensured a fair comparison across all models.}

{\noindent \textbf{Segmentation Optimization:}  
To improve segmentation accuracy, a composite loss function incorporating Cross-Entropy Loss and Dice Loss was applied, ensuring precise boundary detection and mitigating class imbalance. Data augmentation included Random Rotation, Scaling, Horizontal Flipping, and Gaussian Noise Injection. In MobileSAM-based segmentation, bounding box predictions from YOLOv10 served as prompts for refining instance segmentation, improving object localization accuracy. Evaluation metrics included {mean Intersection-over-Union (mIoU)}, {mean Pixel-Level Accuracy (mPLA)}, inference time, number of parameters, and memory usage.}

\subsection{Loss Function for Object Detection}
{\noindent \textbf{Cost Function and Loss Terms:}  
The total training loss was formulated as:
\begin{equation}
\mathcal{L} = \lambda_{iou} L_{iou} + \lambda_{cls} L_{cls} + \lambda_{obj} L_{obj}
\end{equation}
where \(\lambda_{iou}\), \(\lambda_{cls}\), and \(\lambda_{obj}\) balance the loss components. Bounding box loss (CIoU) improves localization accuracy, classification loss (BCE) optimizes class predictions, and objectness loss enhances confidence estimation. These collectively refine detection performance.}

\subsection{Evaluation Metrics}
Both object detection and segmentation models were evaluated using the following metrics:

\begin{itemize}
    \item \textbf{Precision:}  
    \begin{equation}
    \text{Precision} = \frac{TP}{TP + FP}
    \end{equation}

    \item \textbf{Recall:}  
    \begin{equation}
    \text{Recall} = \frac{TP}{TP + FN}
    \end{equation}

    \item \textbf{F1-Score:}  
    \begin{equation}
    \text{F1-score} = 2 \times \frac{\text{Precision} \times \text{Recall}}{\text{Precision} + \text{Recall}}
    \end{equation}

    \item \textbf{Mean Average Precision (mAP):}  
    Evaluated at IoU thresholds of 0.5, 0.75, and 0.95.

    \item \textbf{Mean Pixel-Level Accuracy (mPLA):}  
    \begin{equation}
    \text{mPLA} = \frac{1}{k+1} \sum_{i=0}^{k} \frac{p_{ii}}{\sum_{j=0}^{k} p_{ij}}
    \end{equation}

    \item \textbf{Mean Intersection-over-Union (mIoU):}  
    \begin{equation}
    \text{mIoU} = \frac{1}{k+1} \sum_{i=0}^{k} \frac{p_{ii}}{\sum_{j=0}^{k} (p_{ij} + p_{ji}) - p_{ii}}
    \end{equation}
\end{itemize}

This experimental setup ensured a comprehensive evaluation framework, balancing accuracy, efficiency, and real-time feasibility for deployment on edge AI devices.

\section{Results}
\subsection{Object Detection}
The YOLOv10 model stands out as the most effective for object detection, achieving an impressive mAP50 score of {0.9627}. It excels across other metrics, including mAP75 ({0.7731}) and mAP95 ({0.7178}), along with high precision ({0.9042}), recall ({0.9204}), and F1-score ({0.9123}), as detailed in Table~\ref{tab:det_metrics}. YOLOv10 also boasts an efficient inference time of {0.0437} seconds per image, making it highly suitable for real-time applications.

In addition to its strong detection performance, YOLOv10 demonstrates superior computational efficiency with the smallest number of parameters ({2.776M}) and the lowest memory usage ({10.59MB}) among the compared models (Table~\ref{tab:det_metrics} and Table~\ref{tab:det_time_space}). It maintains its edge in speed, being the fastest with an inference time of just {43.7 ms/image}. In contrast, models such as FasterRCNN-ResNet-50 and YOLOv5, while offering competitive detection performance, are less suitable for real-time applications due to their higher computational demands. For instance, FasterRCNN-ResNet-50 has a significantly higher parameter count ({41.755M}) and memory usage ({159.28MB}) with a slower inference time of {2.3628} seconds. YOLOv5, though faster than FasterRCNN, still requires {0.1155} seconds per image and consumes {203.04MB} of memory, making it more resource-intensive.

\begin{figure*}[!t]
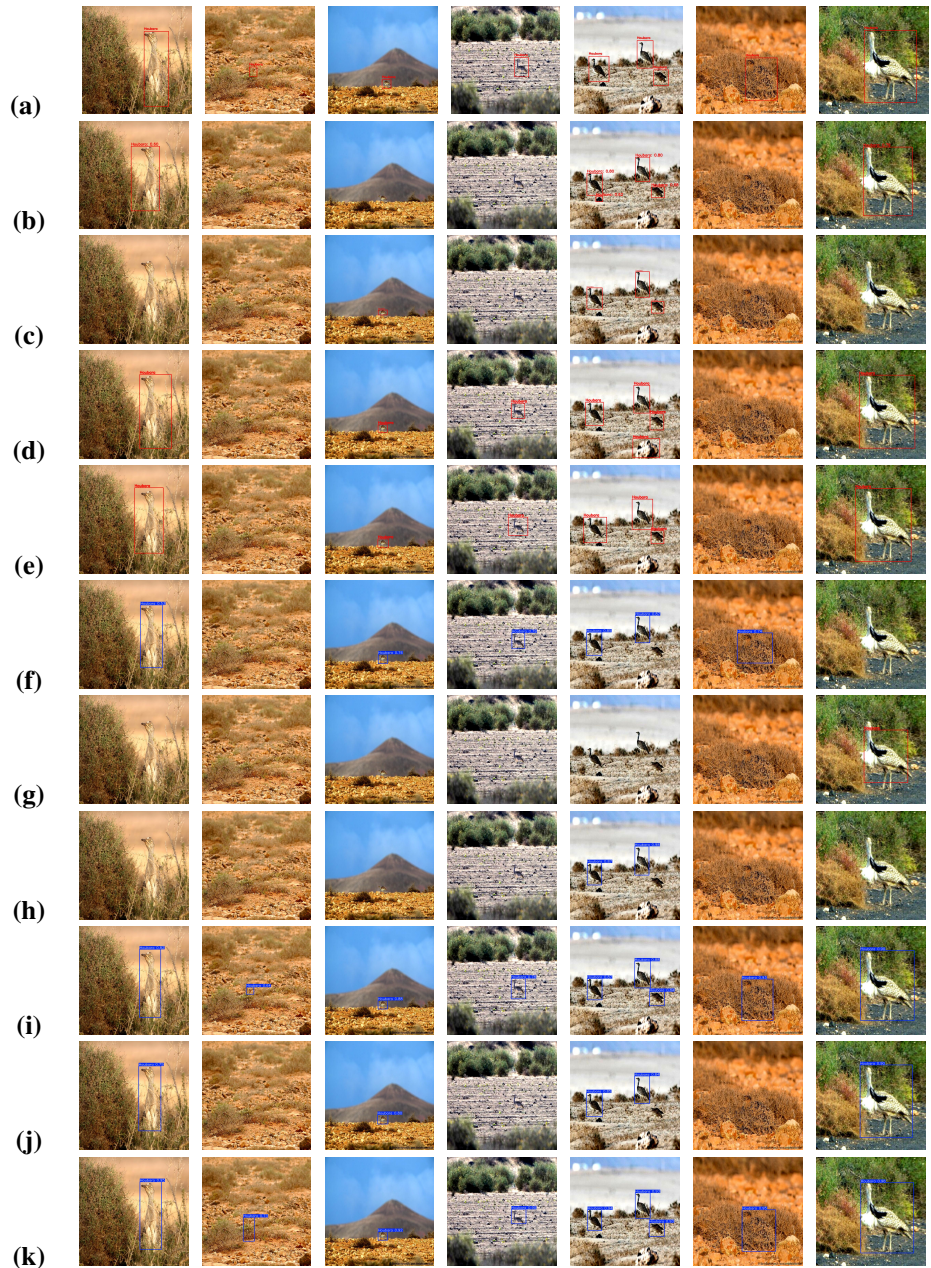

    \centering

    % First row: GT images 1-7
    \begin{minipage}[b]{0.05\textwidth}
        \centering
        \textbf{(a)}
    \end{minipage}%    
    \hspace{-0.02\textwidth} % Adjust this value as needed
    \begin{minipage}[b]{0.95\textwidth}
        \centering
        \foreach \i in {1, 2, 3, 4, 5, 6, 10} {
            \begin{subfigure}{0.1151\textwidth}
                \includegraphics[width=\textwidth]{Detection/GT_\i.jpg}
            \end{subfigure}%
            \hspace{0.01cm}
        }
    \end{minipage}
    
    % Second row: EfficientDet
    \begin{minipage}[b]{0.05\textwidth}
        \centering
        \textbf{(b)}
    \end{minipage}%     
    \hspace{-0.02\textwidth}% Adjust this value as needed
    \begin{minipage}[b]{0.95\textwidth}
        \centering
        \foreach \i in {1, 2, 3, 4, 5, 6, 10} {
            \begin{subfigure}{0.1151\textwidth}
                \includegraphics[width=\textwidth]{Detection/EffiicientDet_\i.jpg}
            \end{subfigure}%
            \hspace{0.01cm}
        }
    \end{minipage}
    
    % Third row: FCOS
    \begin{minipage}[b]{0.05\textwidth}
        \centering
        \textbf{(c)}
    \end{minipage}%    
    \hspace{-0.02\textwidth}% Adjust this value as needed
    \begin{minipage}[b]{0.95\textwidth}
        \centering
        \foreach \i in {1, 2, 3, 4, 5, 6, 10} {
            \begin{subfigure}{0.1151\textwidth}
                \includegraphics[width=\textwidth]{Detection/FCOS_\i.jpg}
            \end{subfigure}%
            \hspace{0.01cm}
        }
    \end{minipage}
    
    % Fourth row: Faster R-CNN (FRCNN)
    \begin{minipage}[b]{0.05\textwidth}
        \centering
        \textbf{(d)}
    \end{minipage}%     
    \hspace{-0.02\textwidth}% Adjust this value as needed
    \begin{minipage}[b]{0.95\textwidth}
        \centering
        \foreach \i in {1, 2, 3, 4, 5, 6, 10} {
            \begin{subfigure}{0.1151\textwidth}
                \includegraphics[width=\textwidth]{Detection/FRCNN_\i.jpg}
            \end{subfigure}%
            \hspace{0.01cm}
        }
    \end{minipage}
    
    % Fifth row: RetinaNet
    \begin{minipage}[b]{0.05\textwidth}
        \centering
        \textbf{(e)}
    \end{minipage}%     
    \hspace{-0.02\textwidth}% Adjust this value as needed
    \begin{minipage}[b]{0.95\textwidth}
        \centering
        \foreach \i in {1, 2, 3, 4, 5, 6, 10} {
            \begin{subfigure}{0.1151\textwidth}
                \includegraphics[width=\textwidth]{Detection/Retina_\i.jpg}
            \end{subfigure}%
            \hspace{0.01cm}
        }
    \end{minipage}
    
    % Sixth row: RTDETR
    \begin{minipage}[b]{0.05\textwidth}
        \centering
        \textbf{(f)}
    \end{minipage}%     
    \hspace{-0.02\textwidth}% Adjust this value as needed
    \begin{minipage}[b]{0.95\textwidth}
        \centering
        \foreach \i in {1, 2, 3, 4, 5, 6, 10} {
            \begin{subfigure}{0.1151\textwidth}
                \includegraphics[width=\textwidth]{Detection/RTDETR_\i.jpg}
            \end{subfigure}%
            \hspace{0.01cm}
        }
    \end{minipage}
    
    % Seventh row: SSD
    \begin{minipage}[b]{0.05\textwidth}
        \centering
        \textbf{(g)}
    \end{minipage}%     
    \hspace{-0.02\textwidth}% Adjust this value as needed
    \begin{minipage}[b]{0.95\textwidth}
        \centering
        \foreach \i in {1, 2, 3, 4, 5, 6, 10} {
            \begin{subfigure}{0.1151\textwidth}
                \includegraphics[width=\textwidth]{Detection/SSD_\i.jpg}
            \end{subfigure}%
            \hspace{0.01cm}
        }
    \end{minipage}
    
    % Eighth row: YOLOv5
    \begin{minipage}[b]{0.05\textwidth}
        \centering
        \textbf{(h)}
    \end{minipage}%     
    \hspace{-0.02\textwidth}% Adjust this value as needed
    \begin{minipage}[b]{0.95\textwidth}
        \centering
        \foreach \i in {1, 2, 3, 4, 5, 6, 10} {
            \begin{subfigure}{0.1151\textwidth}
                \includegraphics[width=\textwidth]{Detection/yolo5_\i.jpg}
            \end{subfigure}%
            \hspace{0.01cm}
        }
    \end{minipage}
    
    % Ninth row: YOLOv8
    \begin{minipage}[b]{0.05\textwidth}
        \centering
        \textbf{(i)}
    \end{minipage}%    
    \hspace{-0.02\textwidth}% Adjust this value as needed
    \begin{minipage}[b]{0.95\textwidth}
        \centering
        \foreach \i in {1, 2, 3, 4, 5, 6, 10} {
            \begin{subfigure}{0.1151\textwidth}
                \includegraphics[width=\textwidth]{Detection/yolo8_\i.jpg}
            \end{subfigure}%
            \hspace{0.01cm}
        }
    \end{minipage}
    
    % Tenth row: YOLOv9
    \begin{minipage}[b]{0.05\textwidth}
        \centering
        \textbf{(j)}
    \end{minipage}%     
    \hspace{-0.02\textwidth}% Adjust this value as needed
    \begin{minipage}[b]{0.95\textwidth}
        \centering
        \foreach \i in {1, 2, 3, 4, 5, 6, 10} {
            \begin{subfigure}{0.1151\textwidth}
                \includegraphics[width=\textwidth]{Detection/yolo9_\i.jpg}
            \end{subfigure}%
            \hspace{0.01cm}
        }
    \end{minipage}
    
    % Eleventh row: YOLOv10
    \begin{minipage}[b]{0.05\textwidth}
        \centering
        \textbf{(k)}
    \end{minipage}%     
    \hspace{-0.02\textwidth}% Adjust this value as needed
    \begin{minipage}[b]{0.95\textwidth}
        \centering
        \foreach \i in {1, 2, 3, 4, 5, 6, 10} {
            \begin{subfigure}{0.1151\textwidth}
                \includegraphics[width=\textwidth]{Detection/yolo10_\i.jpg}
            \end{subfigure}%
            \hspace{0.01cm}
        }
    \end{minipage}

    \caption{Comparison of object detection results across different models. (a) Ground Truth (GT), (b) EfficientDet, (c) FCOS, (d) Faster R-CNN (FRCNN), (e) RetinaNet, (f) RTDETR, (g) SSD, (h) YOLOv5, (i) YOLOv8, (j) YOLOv9, (k) YOLOv10.}
    \label{fig:dect_compare}
\end{figure*}

Figure~\ref{fig:dect_compare} illustrates the performance of various object detection models. Here are some observations and additional suggestions for improvement:

\begin{itemize}
  
    \item \textbf{EfficientDet, FCOS, Faster R-CNN (FRCNN), RetinaNet, RTDETR, SSD, YOLOv5, and YOLOv8:} These models exhibit similar challenges, particularly in detecting Houbara bustards that are hidden, camouflaged, or occluded. They tend to miss instances when multiple Houbara bustards are present or when they are positioned at a distance. For instance, these models often fail to detect Houbara bustards that are partially obscured or blending into the background, resulting in missed detections in complex scenes.
    
    \item \textbf{YOLOv9:} Shows improved performance over the aforementioned other models, handling multi-instance scenes better and providing more reliable detection of Houbara bustards. Despite these improvements, it still faces some difficulties with highly camouflaged or distant instances.

    \item \textbf{YOLOv10:} Demonstrates remarkable capabilities in detecting Houbara bustards under a variety of conditions, including challenging scenarios such as low-light environments and occlusion. YOLOv10 effectively identifies hidden or camouflaged Houbara bustards and manages multiple instances within a scene with high accuracy. Its superior detection and localization performance are evident across all evaluated conditions, setting it apart from other models.

\end{itemize}

These insights and additional suggestions highlight YOLOv10's exceptional performance and its importance for effective and real-time wildlife monitoring.

\begin{table}[t]
\caption{Performance Metrics of Object Detection Models}
\label{tab:det_metrics}
\centering
\begin{tabular}{lcccccc}
\hline
\textbf{Model Name} & \textbf{mAP50} $\uparrow$ & \textbf{mAP75}  $\uparrow$  & \textbf{mAP95}  $\uparrow$ & \textbf{Precision}  $\uparrow$ & \textbf{Recall}  $\uparrow$ & \textbf{F1-score} $\uparrow$   \\
\hline
RetinaNet & 0.9433 & 0.6575 & 0.4025 & 0.8615 & 0.8411 & 0.8510  \\
FCOS & 0.6806 & 0.4054 & 0.3422 & 0.6985 & 0.7176 & 0.7081 \\
FasterRCNN-ResNet-50 & 0.8291 & 0.6234 & 0.4907 & 0.7649 & 0.7858 & 0.7753 \\
FasterRCNN-MobileNetV3 & 0.9526 & 0.7430 & 0.4178 & 0.8741 & 0.8979 & 0.8860 \\
RTDETR & 0.9255 & 0.6693 & 0.5685 & 0.8447 & 0.8844 & 0.8644 \\
MobileNetSSD & 0.8433 & 0.5183 & 0.4471 & 0.6560 & 0.5777 & 0.6138 \\
EfficientDet & 0.7410 & 0.5239 & 0.6014 & 0.6949 & 0.8698 & 0.7425 \\ YOLOv5  & 0.9672 & 0.7286 & 0.6855 & 0.9356 & 0.9365 & 0.9360 \\ YOLOv8  & 0.9692 & 0.7367 & 0.6897 & 0.9373 & 0.9345 & 0.9359 \\ YOLOv9  & 0.9693 & 0.7640 & 0.7107 & 0.9300 & 0.9364 & 0.9332 \\ YOLOv10 & \textbf{0.9627} & \textbf{0.7731} & \textbf{0.7178} & \textbf{0.9042} & \textbf{0.9204} & \textbf{0.9123} \\
\hline
\end{tabular}
\end{table}

\begin{table}[t]
\centering
\caption{ Computational Performance Metrics for Object Detection Models}
\begin{tabular}{lccc}
\hline
\textbf{Model Name} & \textbf{Time (Sec.)} $\downarrow$ & \textbf{No of Par. (M)} $\downarrow$ & \textbf{Memory (MB)} $\downarrow$ \\
\hline
RetinaNet & 0.0761 & ~34.015 & 129.76 \\
FCOS & 1.4801 & ~32.270 & 123.10 \\

FasterRCNN-ResNet-50 & 2.3628 & ~41.755 & 159.28 \\
FasterRCNN-MobileNetV3 & 0.176 & ~19.386 & 73.95 \\
RTDETR & 0.0135 & ~32.970 & 125.77 \\
MobileNetSSD & 0.1731 & ~3.440 & 13.12 \\
EfficientDet & 0.1151 & ~6.626 & 25.28 \\
YOLOv5 & 0.1155 & ~53.225 & 203.04 \\
YOLOv8 & 0.0739 & ~43.692 & 166.67 \\
YOLOv9 & 0.0568 & ~25.591 & 97.62 \\
YOLOv10 & \textbf{0.0437} & \textbf{~2.776} & \textbf{10.59} \\
\hline
\end{tabular}
\label{tab:det_time_space}
\end{table}

\subsection{Segmentation}

\begin{figure*}[!t]
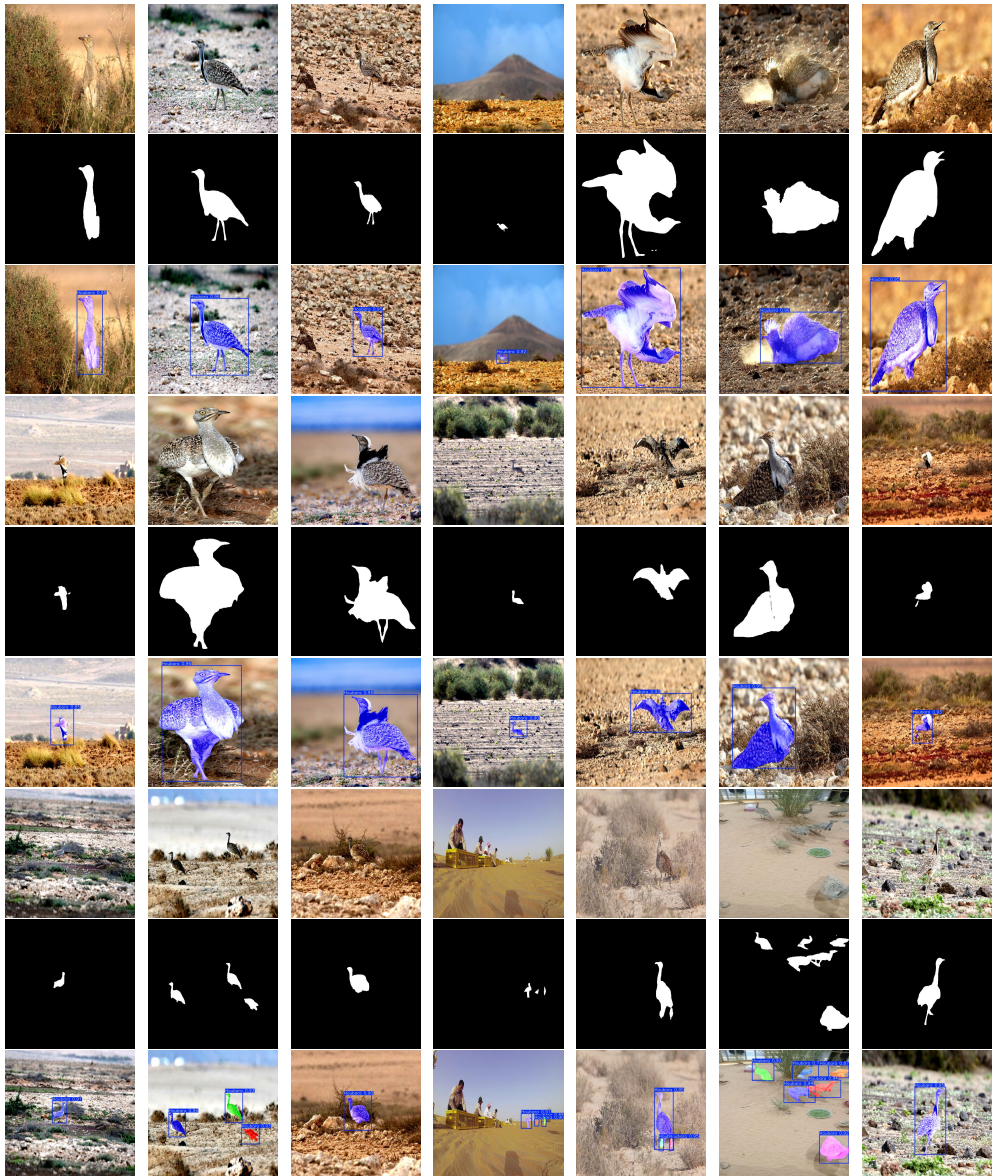

    \centering
    
    % First row: GT images 1-6
    \begin{subfigure}[b]{1.0\textwidth}
        \centering
        \foreach \i in {1, 2, 3, 4, 5, 6, 7} {
            \begin{subfigure}{0.13\textwidth}
                \includegraphics[width=\textwidth]{Seg_det/g\i.jpg}
            \end{subfigure}%
            \hspace{0.01cm}
        }
    \end{subfigure}
     % Third row: Annotated images 1-6
    \begin{subfigure}[b]{1.0\textwidth}
        \centering
        \foreach \i in {1, 2, 3, 4, 5, 6, 7} {
            \begin{subfigure}{0.13\textwidth}
                \includegraphics[width=\textwidth]{Seg_det/mask\i.jpg}
            \end{subfigure}%
            \hspace{0.01cm}
        }
    \end{subfigure}
    \begin{subfigure}[b]{1.0\textwidth}
        \centering
        \foreach \i in {1, 2, 3, 4, 5, 6, 7} {
            \begin{subfigure}{0.13\textwidth}
                \includegraphics[width=\textwidth]{Seg_det/\i.jpg}
            \end{subfigure}%
            \hspace{0.01cm}
        }
    \end{subfigure}
    
     % First row: GT images 1-6
    \begin{subfigure}[b]{1.0\textwidth}
        \centering
        \foreach \i in {8, 9, 10, 11, 12, 13, 14} {
            \begin{subfigure}{0.13\textwidth}
                \includegraphics[width=\textwidth]{Seg_det/g\i.jpg}
            \end{subfigure}%
            \hspace{0.01cm}
        }
    \end{subfigure}
     % Third row: Annotated images 1-6
    \begin{subfigure}[b]{1.0\textwidth}
        \centering
        \foreach \i in {8, 9, 10, 11, 12, 13, 14} {
            \begin{subfigure}{0.13\textwidth}
                \includegraphics[width=\textwidth]{Seg_det/mask\i.jpg}
            \end{subfigure}%
            \hspace{0.01cm}
        }
    \end{subfigure}
    \begin{subfigure}[b]{1.0\textwidth}
        \centering
        \foreach \i in {8, 9, 10, 11, 12, 13, 14} {
            \begin{subfigure}{0.13\textwidth}
                \includegraphics[width=\textwidth]{Seg_det/\i.jpg}
            \end{subfigure}%
            \hspace{0.01cm}
        }
    \end{subfigure}

     % First row: GT images 1-6
    \begin{subfigure}[b]{1.0\textwidth}
        \centering
        \foreach \i in {15, 16, 17, 18, 19, 20, 21} {
            \begin{subfigure}{0.13\textwidth}
                \includegraphics[width=\textwidth]{Seg_det/g\i.jpg}
            \end{subfigure}%
            \hspace{0.01cm}
        }
    \end{subfigure}
     % Third row: Annotated images 1-6
    \begin{subfigure}[b]{1.0\textwidth}
        \centering
        \foreach \i in {15, 16, 17, 18, 19, 20, 21} {
            \begin{subfigure}{0.13\textwidth}
                \includegraphics[width=\textwidth]{Seg_det/mask\i.jpg}
            \end{subfigure}%
            \hspace{0.01cm}
        }
    \end{subfigure}
    \begin{subfigure}[b]{1.0\textwidth}
        \centering
        \foreach \i in {15, 16, 17, 18, 19, 20, 21} {
            \begin{subfigure}{0.13\textwidth}
                \includegraphics[width=\textwidth]{Seg_det/\i.jpg}
            \end{subfigure}%
            \hspace{0.01cm}
        }
    \end{subfigure}

   \caption{Real-time Houbara analysis: (a) Input images (first, fourth, and seventh rows). (b) Predicted instance segmentation masks by MobileSAM (second, fifth, and eighth rows). (c) Combined view: Input images with object detections (bounding boxes) and MobileSAM segmentation masks (third, sixth, and ninth rows).}
    \label{fig:gt_mask_bb}
\end{figure*}

The combination of YOLOv10 and MobileSAM proves to be superior for segmentation, achieving the highest mPLA of 0.9773 and an mIoU of 0.7421 (see Table~\ref{tab:segmentation_results}). Figure~\ref{fig:gt_mask_bb} provides a qualitative comparison, showing input images (panel a), predicted instance segmentation masks by MobileSAM (panel b), and the combined view with object detections and segmentation masks (panel c).

The YOLOv10 + MobileSAM combination not only detects objects accurately but also segments the regions of interest (RoI) with high precision. The mPLA score reflects the model's overall ability to correctly classify each pixel, while the mIoU metric evaluates how well the predicted segmentation matches the ground truth. These results demonstrate a significant improvement over previous models, such as YOLOv9 + MobileSAM and DeepLabV3.

MobileSAM's lightweight architecture enables rapid inference with a time of 0.1075 seconds per image, further supporting its deployment on mobile and edge devices. This speed ensures that the system can perform both object detection and segmentation in real-time, making it feasible for practical applications such as monitoring and analyzing Houbara behavior in the wild. The efficient processing time of YOLOv10 + MobileSAM ensures not only high accuracy but also real-time deployment, making it ideal for on-the-go fieldwork where immediate analysis is crucial.

Figure~\ref{fig:gt_mask_bb} provides a qualitative comparison of the segmentation results across different models, showcasing the input images (panel a), predicted instance segmentation masks generated by MobileSAM (panel b), and the combined view with both object detections and segmentation masks overlaid (panel c). Once more, the YOLOv10 + MobileSAM combination performs particularly well, accurately detecting objects and segmenting the RoI with high precision. As seen in panel (c), the segmentation masks align closely with the detected objects, confirming the high mPLA (0.9773) and mIoU (0.7421) scores.

\begin{figure*}[!t]
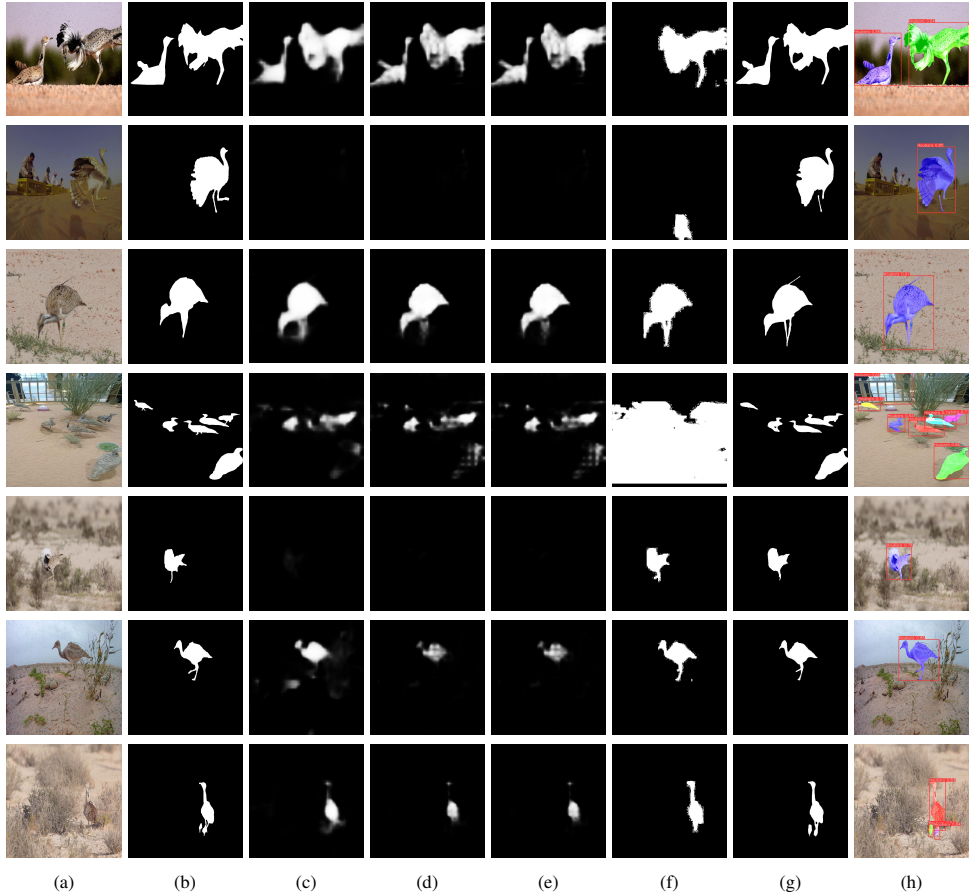

    \centering
    % First column: Input images (a)
    \begin{subfigure}[b]{0.1151\textwidth}
        \centering
        \foreach \i in {1, 2, 3, 4, 5, 6, 7} {
            \includegraphics[width=\textwidth]{Compare/input_\i.jpg}\\
            \vspace{0.1cm}
        }
        \caption*{(a)}
    \end{subfigure}
    % Second column: Ground Truth (b)
    \begin{subfigure}[b]{0.1151\textwidth}
        \centering
        \foreach \i in {1, 2, 3, 4, 5, 6, 7} {
            \includegraphics[width=\textwidth]{Compare/mask_\i.jpg}\\
            \vspace{0.1cm}
        }
        \caption*{(b)}
    \end{subfigure}
    % Third column: DeepLabV3 (c)
    \begin{subfigure}[b]{0.1151\textwidth}
        \centering
        \foreach \i in {1, 2, 3, 4, 5, 6, 7} {
            \includegraphics[width=\textwidth]{Compare/DeepLabV3_\i.jpg}\\
            \vspace{0.1cm}
        }
        \caption*{(c)}
    \end{subfigure}
    % Fourth column: FCN (d)
    \begin{subfigure}[b]{0.1151\textwidth}
        \centering
        \foreach \i in {1, 2, 3, 4, 5, 6, 7} {
            \includegraphics[width=\textwidth]{Compare/FCN_\i.jpg}\\
            \vspace{0.1cm}
        }
        \caption*{(d)}
    \end{subfigure}
    % Fifth column: LRASPP (e)
    \begin{subfigure}[b]{0.1151\textwidth}
        \centering
        \foreach \i in {1, 2, 3, 4, 5, 6, 7} {
            \includegraphics[width=\textwidth]{Compare/LRASPP_\i.jpg}\\
            \vspace{0.1cm}
        }
        \caption*{(e)}
    \end{subfigure}
    % Sixth column: MobileSAM Baseline (f)
    \begin{subfigure}[b]{0.1151\textwidth}
        \centering
        \foreach \i in {1, 2, 3, 4, 5, 6, 7} {
            \includegraphics[width=\textwidth]{Compare/MobileSAMFull_\i.jpg}\\
            \vspace{0.1cm}
        }
        \caption*{(f)}
    \end{subfigure}
    % Seventh column: YOLOv9+MobileSAM (g)
    \begin{subfigure}[b]{0.1151\textwidth}
        \centering
        \foreach \i in {1, 2, 3, 4, 5, 6, 7} {
            \includegraphics[width=\textwidth]{Compare/MobileSAM_\i.jpg}\\
            \vspace{0.1cm}
        }
        \caption*{(g)}
    \end{subfigure}
    % Eighth column: YOLOv10+MobileSAM (h)
    \begin{subfigure}[b]{0.1151\textwidth}
        \centering
        \foreach \i in {1, 2, 3, 4, 5, 6, 7} {
            \includegraphics[width=\textwidth]{Compare/Over_\i.jpg}\\
            \vspace{0.1cm}
        }
        \caption*{(h)}
    \end{subfigure}
    
    % General caption
    \caption{\textbf{Qualitative comparison of segmentation results.}
    This figure presents a qualitative comparison of segmentation performance across various models. Each column displays: (a) Input Images, (b) Ground Truth Annotations, (c) DeepLabV3 Results, (d) FCN Results, (e) LRASPP Results, (f) MobileSAM Baseline Results, (g) YOLOv9+MobileSAM Results, and (h) Overlapping Results of YOLOv10+MobileSAM with Input Images. The overlapping panel highlights the superior object localization and segmentation precision achieved by YOLOv10+MobileSAM.}

    \label{fig:segmentation_comparison}
\end{figure*}

Figure~\ref{fig:segmentation_comparison} reveals several key insights into the performance of the models. YOLOv10 + MobileSAM stands out with the highest mPLA and mIoU scores. This indicates superior precision in object localization and segmentation, as this model effectively identifies subtle features and defines object boundaries more accurately compared to other methods. In contrast, DeepLabV3 and FCN exhibit lower accuracy and segmentation quality. Their masks often show blurring and less defined boundaries, which can impact their effectiveness in applications requiring precise segmentation.

\begin{table}[t]
\centering
\caption{Performance Comparison of Segmentation Models}
\begin{tabular}{lccccc}
\hline
\textbf{Model Name} & \textbf{mPLA} $\uparrow$ & \textbf{mIoU} $\uparrow$ & \textbf{Time (Sec.)} $\downarrow$ & \textbf{No. of Par. (M)} $\downarrow$ & \textbf{Memory (MB)} $\downarrow$ \\
\hline
DeepLabV3\_ResNet50 & 0.7729 & 0.3372 & 0.3357 & ~42.004 & 160.23 \\
LRASPP\_MobileNet\_V3\_Large & 0.8353 & 0.3102 & 0.2634 & \textbf{~3.221} & \textbf{12.29} \\
FCN\_ResNet50 & 0.8624 & 0.3890 & 1.6016 & ~35.322 & 134.74 \\
Baseline MobileSAM & 0.8216 & 0.1503 & 0.6285 & ~10.130 & 38.64 \\
YOLOv9 + MobileSAM & 0.9654 & 0.7242 & 0.1292 & ~35.721 & 136.26 \\
\textbf{YOLOv10 + MobileSAM} & \textbf{0.9773} & \textbf{0.7421} & \textbf{0.1075} & ~12.906 & 49.23\\
% \hline
% MobileYOLACT & - & - & - & - & - \\
% SOLO & - & - & - & - & - \\
% BlendMask & - & - & - & - & - \\
% CenterMask & - & - & - & - & - \\
% BlitzMask & - & - & - & - & - \\
\hline
\end{tabular}
\label{tab:segmentation_results}
\end{table}

\section{Discussions}

In this section, we analyze the performance of our proposed models and compare them with existing approaches, focusing on their effectiveness in object detection and segmentation, particularly under challenging conditions.

%%%%%%%%%%%%%%%%%%%%%%%%%%%%%%%%%%%%%%%%%%%%%%%%%%%%%%%%

In this section, we analyze the impact of different image conditions on detection performance. The analysis considers error rates across various challenging scenarios, including occlusion, extreme lighting, and small object detection.

\begin{figure}[h]
    \centering
    \includegraphics[width=0.9\linewidth]{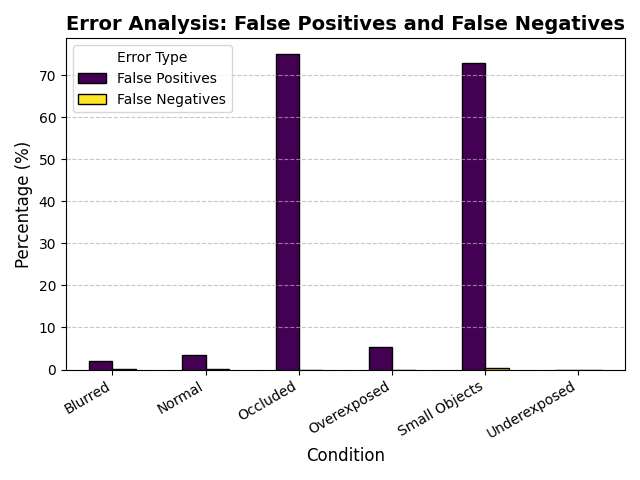}
    \caption{Error analysis of false positives and false negatives across different conditions.}
    \label{fig:error_analysis}
\end{figure}

Figure~\ref{fig:error_analysis} presents an error breakdown by category, where occluded and small-object images exhibit the highest false negative rates (FN), highlighting their impact on detection performance. Additionally, false positives (FP) are relatively high in overexposed conditions, likely due to misclassification in high-brightness environments. 

\begin{figure}[h]
    \centering
    \includegraphics[width=0.9\linewidth]{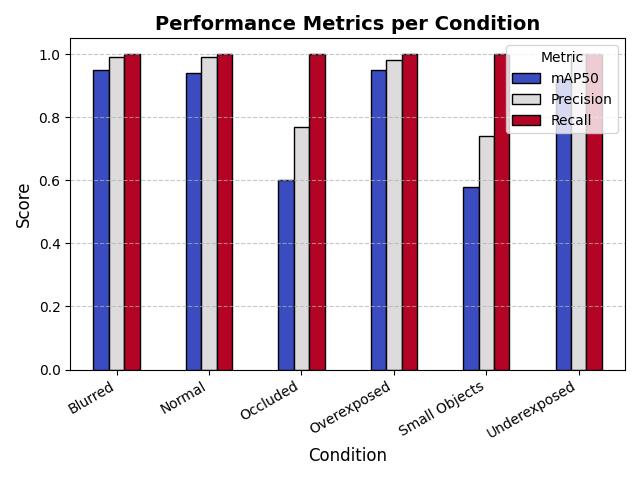}
    \caption{Performance metrics across different image conditions, comparing mAP50, Precision, and Recall.}
    \label{fig:performance_summary}
\end{figure}

Figure~\ref{fig:performance_summary} provides a comparative analysis of mAP50, Precision, and Recall across image categories. While our model maintains strong performance in normal and blurred conditions, occluded and small-object images show a significant drop in recall, confirming that occlusion and object size are key limiting factors.

The outstanding performance of the YOLOv10 model, as demonstrated in Table~\ref{tab:det_metrics}, Table~\ref{tab:det_time_space}, and Figure~\ref{fig:dect_compare}, highlights its effectiveness in complex object detection scenarios. Its high mAP50, mAP75, and mAP95 scores, along with superior precision, recall, and F1-score, confirm its robustness in accurately identifying objects, particularly the Houbara bustard. Furthermore, its fast inference time, minimal parameter count, and low memory usage make it highly suitable for real-time applications, as shown in Table~\ref{tab:det_time_space}.

Alternative models, such as EfficientDet and FCOS, exhibit difficulties in detecting Houbara bustards, especially in occluded or camouflaged settings. EfficientDet's reliance on predefined anchor boxes limits its adaptability to varying object appearances, particularly for cryptic species. FCOS, while employing an anchor-free approach, struggles with detecting small objects across varying scales and complex backgrounds. Likewise, Faster R-CNN (FRCNN) and RetinaNet, despite their strengths, exhibit suboptimal performance in highly occluded or low-contrast scenarios due to their dependence on region proposals and anchor-based methods that may not generalize well under challenging conditions.

YOLOv10’s advanced architecture effectively mitigates these limitations. It consistently demonstrates superior detection and localization capabilities across diverse conditions, outperforming YOLOv9, as depicted in Figure~\ref{fig:dect_compare}. Its improved feature extraction and integration of multi-scale features make it particularly well-suited for detecting camouflaged or partially obscured subjects in wildlife monitoring applications, as illustrated in Figure~\ref{fig:dect_compare}.

Other models, while excelling in specific aspects, often lack real-time processing efficiency or struggle in scenarios requiring both detection and segmentation. For instance, Faster R-CNN and YOLOv5, though competent in object detection, do not provide seamless segmentation, leading to suboptimal results in applications that demand precise boundary delineation, as demonstrated in Figure~\ref{fig:segmentation_comparison}.

Our proposed YOLOv10 + MobileSAM framework significantly enhances performance by leveraging the complementary strengths of both models. YOLOv10’s high detection accuracy is augmented by MobileSAM’s advanced segmentation capabilities, enabling precise object delineation even under challenging conditions. This combination results in improved segmentation accuracy and overall performance, as detailed in Table~\ref{tab:segmentation_results} and visually depicted in Figure~\ref{fig:segmentation_comparison}.

MobileSAM further enhances YOLOv10’s capabilities by delivering highly accurate segmentation, effectively resolving challenges related to occlusion and cryptic object appearances. This combined approach addresses limitations observed in other models, which may struggle to integrate detection and segmentation seamlessly, as demonstrated in Table~\ref{tab:segmentation_results}.

{Unlike prior YOLO + SAM integrations, our approach incorporates a Threading Detection Model (TDM), which enables parallelized inference, significantly reducing computation time and making our framework highly suitable for mobile wildlife monitoring applications. While conventional detection and segmentation pipelines operate sequentially, our method achieves substantial performance gains by processing detection and segmentation concurrently. This optimization leads to a notable improvement in real-time performance, with YOLOv10 achieving a mean average precision (mAP) of {0.9627} and MobileSAM attaining an Intersection-over-Union (IoU) of {0.7421}.}

{One of the key advancements introduced by TDM is the significant reduction in processing latency. Without TDM, the system experienced an average inference delay of approximately {5 seconds per frame}, which posed a major limitation for real-time mobile robotic applications such as HuBot~\cite{SAADSAOUD2025102939}. This latency hindered timely decision-making, particularly in dynamic environments where rapid response is crucial for navigation and target tracking. By integrating TDM, our system achieves real-time performance with significantly lower inference latency, ensuring enhanced responsiveness in real-world deployments.}

{Reducing inference latency is critical for applications such as autonomous robotic systems and real-time species monitoring, where delays in object detection and segmentation can lead to incorrect decisions or missed observations. The optimizations introduced by TDM enable efficient operation in real-time conservation applications, ensuring immediate feedback for mobile robots tracking elusive species in complex terrains. A comparative analysis showcasing the system’s performance with and without TDM is available in our GitHub repository.}

In summary, YOLOv10 surpasses other models in both detection accuracy and computational efficiency. The incorporation of MobileSAM further strengthens its segmentation performance, making it an ideal solution for wildlife monitoring applications that require both real-time detection and precise segmentation. The practical advantages of this integrated approach underscore its potential for broader applications in real-time conservation monitoring and mobile robotics.

\section{Limitations and Future Work}
\label{sec:limitations}

Despite the high performance of YOLOv10 and MobileSAM in object detection and segmentation tasks, several limitations remain. Our {\textbf{error analysis (Figure~\ref{fig:error_analysis})}} highlights the {\textbf{challenges posed by occlusion and small object detection}}, which significantly affect model recall. These findings suggest areas where further improvements are needed.

\subsection{Challenges}

\begin{itemize}
    \item {\textbf{Highly cryptic objects:} YOLOv10 and MobileSAM, like other detection models, can struggle with objects that blend seamlessly into their surroundings. These challenges are particularly evident in occluded and camouflaged images, where false negatives (FN) remain high.}

    \item {\textbf{Inference time for high-resolution images:} While YOLOv10 and MobileSAM are relatively efficient compared to other frameworks, processing high-resolution images can increase inference time, potentially hindering real-time performance on devices with limited computational resources.}

    \item {\textbf{Impact of occlusion and small object size:} Our quantitative evaluation (Figure~\ref{fig:performance_summary}) shows that occlusion leads to a significant drop in recall (0.62) and an increased FN rate. Similarly, small object detection remains a challenge, impacting overall accuracy.}

    \item {\textbf{Model generalization:} The models, particularly YOLOv10 and MobileSAM, are trained on a specific dataset (Houbara bustard) and may not generalize well to other types of wildlife or environmental conditions. This could impact performance in varied or unforeseen scenarios.}

    \item {\textbf{Scalability:} The approach demonstrated with YOLOv10 and MobileSAM, while effective, may not scale seamlessly to other object detection or segmentation tasks without additional tuning or adjustments. Adapting the models for different applications may require substantial re-training or fine-tuning efforts.}

\end{itemize}

\subsection{Future Work}

To address these limitations, several directions for future research and development can be explored:

\begin{itemize}
    \item {\textbf{Enhanced camouflage detection:} Developing advanced algorithms or incorporating additional features such as texture analysis and multi-spectral imaging could improve the detection of cryptic animals. Leveraging transformer-based architectures may also help in learning more discriminative features for background separation.}
    
    \item {\textbf{Optimization for high-resolution images:} Future work could focus on optimizing YOLOv10 and MobileSAM for processing high-resolution images more efficiently. Techniques such as image tiling, adaptive resizing, or hardware acceleration could be investigated to reduce inference time while maintaining high accuracy.}
    
    \item {\textbf{Evaluating Generalization on Additional Datasets:} While this study focuses on the Houbara bustard dataset, future work will explore the application of our approach to publicly available bird datasets such as CUB-200 and NABirds. This will provide further validation of our method’s robustness and adaptability to different species and ecological environments.}

    \item {\textbf{Cross-domain generalization:} To improve the models’ generalization capabilities, research could be directed towards domain adaptation techniques. This involves adapting the models trained on one dataset to perform well on another dataset with different characteristics, potentially through methods such as domain adversarial training, self-supervised learning, or data augmentation.}

    \item {\textbf{Integration with Multi-Sensor Systems:} Combining our detection model with thermal, infrared, or depth sensors could enhance object detection under challenging conditions such as low-light or occlusion-heavy environments. This would extend the applicability of the model to nocturnal species monitoring and adverse weather conditions.}

    \item {\textbf{Super-resolution for small object detection:} Small objects present a major challenge, as seen in Figure~\ref{fig:error_analysis}. Future work will investigate super-resolution techniques to enhance feature extraction for tiny object detection, particularly when working with high-altitude or aerial imagery.}

    \item {\textbf{Scalability and adaptation:} Investigating methods to scale the current approach to other object detection and segmentation tasks will be important. This could involve exploring modular learning architectures or transfer learning-based approaches to quickly adapt the models to new domains with minimal retraining.}
    
    \item {\textbf{Real-Time Deployment Enhancements:} While the proposed method achieves real-time performance, further optimizations can be made to reduce computational overhead. Exploring lightweight model architectures, model pruning, and quantization techniques could improve efficiency for deployment on edge devices with limited processing power.}

\end{itemize}

\section{Conclusion}
\label{sec:conclusion}

This study introduces a two-stage deep learning framework for real-time Houbara bird detection and segmentation on mobile platforms. The Threading Detection Model (TDM) enables concurrent processing of YOLOv10 for detection and MobileSAM for segmentation, significantly reducing inference delays and optimizing real-time performance. This allows for more efficient tracking of Houbara birds, overcoming challenges posed by its camouflage and mobile device constraints. Our fine-tuned YOLOv10 model achieves a mAP50 of 0.9627, mAP75 of 0.7731, and mAP95 of 0.7178, demonstrating superior performance over existing state-of-the-art methods, while MobileSAM segmentation ensures high accuracy with an mIoU of 0.7421. Additionally, the introduction of a 40,000-image Houbara dataset enhances model training, evaluation, and conservation research. By eliminating the limitations of traditional monitoring methods, our framework provides a robust tool for real-time wildlife monitoring, supporting conservation efforts by improving animal tracking and behavioural understanding. This work thus contributes to advancing mobile AI applications in conservation, with potential for broader adaptation to other cryptic species and resource-limited environments.

% \section*{Data Availability Statement:}

% The code and dataset used in this study are publicly available on GitHub at https://github.com/LyesSaadSaoud/mobile-houbara-detseg. The dataset will be available on Zenodo within a reasonable timeframe. For demos and attractive ...: https://lyessaadsaoud.github.io/Threaded-YOLO-SAM-Houbara/.
\section*{Data Availability Statement}

The code and dataset used in this study are publicly available on GitHub at \href{https://github.com/LyesSaadSaoud/mobile-houbara-detseg}{GitHub Repository}. The dataset will also be made available on Zenodo within a reasonable timeframe. For interactive demonstrations and additional resources, please visit: \href{https://lyessaadsaoud.github.io/LyesSaadSaoud-Threaded-YOLO-SAM-Houbara}{Project Website}.

% \section*{Acknowledgement}
% This work is supported by Khalifa University under award: “Khalifa University Center for Autonomous and Robotic Systems RC1-2018-KUCARS” and project “Hubot: Houbara robot for behavioral studies in the field and sampling 8434000544”. Additional Funds and samples used in this study were provided by the International Fund for Houbara Conservation (IFHC). We are grateful to His Highness Sheikh Mohamed bin Zayed Al Nahyan, President of the United Arab Emirates and founder of the IFHC, His Highness Sheikh Theyab bin Mohamed Al Nahyan, Chairman of the IFHC, and His Excellency Mohammed Ahmed Al Bowardi, Deputy Chairman, for their support. This study was conducted in collaboration with Reneco International Wildlife Consultants LLC, a consulting company that manages the IFHC’s conservation programs. We thank Dr Frederic Lacroix, Managing Director of Reneco, for his supervision, as well as all staff of Reneco who participated in data collection.

\section*{Declaration}

 The authors declare that they have no conflicts of interest regarding this manuscript.
\bibliography{sn-bibliography}% common bib file

\end{document}